\documentclass{article}

\usepackage[final]{neurips_2019}

\usepackage[utf8]{inputenc} % allow utf-8 input
\usepackage[T1]{fontenc}    % use 8-bit T1 fonts
\usepackage[hidelinks]{hyperref}       % hyperlinks
\usepackage{url}            % simple URL typesetting
\usepackage{booktabs}       % professional-quality tables
\usepackage{amsfonts}       % blackboard math symbols
\usepackage{nicefrac}       % compact symbols for 1/2, etc.
\usepackage{microtype}      % microtypography

\usepackage{xspace}

\usepackage{times}
\usepackage{epsfig}
\usepackage{graphicx}
\usepackage{amsmath}
\usepackage{amssymb}
\usepackage{float}
\usepackage{subfig}
\usepackage{svg}
\usepackage{multirow}
\usepackage{sidecap}
\sidecaptionvpos{figure}{t}
\usepackage{array}
\usepackage[labelfont=bf]{caption}
\usepackage{bbm}
\usepackage{bm}
\usepackage{wrapfig}
\usepackage[parfill]{parskip}
\makeatletter
\DeclareRobustCommand\onedot{\futurelet\@let@token\@onedot}
\def\@onedot{\ifx\@let@token.\else.\null\fi\xspace}

\def\eg{\emph{e.g}\onedot} 

\def\ie{\emph{i.e}\onedot}

\makeatother

\newcommand{\FIG}{Fig.~}

\newcommand{\EQN}{Eq.~}

\newcommand{\TAB}{Table~}
\newcommand{\SUPMAT}{supplementary material}

\newcommand{\Tcal}{\mathcal{T}}
\newcommand{\Ncal}{\mathcal{N}}
\bmdefine\btheta{\theta}
\bmdefine\bgamma{\gamma}
\bmdefine\bmu{\mu}
\bmdefine\bsigma{\sigma}

\newcommand{\Xcal}{\mathcal{X}}

\def\x{{\boldsymbol x}}
\def\z{{\boldsymbol z}}
\renewcommand{\vec}[1]{\mathbf{#1}}
\newcommand{\mat}[1]{\mathbf{#1}}

\DeclareCaptionLabelFormat{andtable}{#1~#2  \&  \tablename~\thetable}

\let\cite\citep % use citep whenever we call cite

\title{Explicit Disentanglement of Appearance and Perspective in Generative Models}

\makeatletter
\newcommand{\printfnsymbol}[1]{%
  \textsuperscript{\@fnsymbol{#1}}%
}
\makeatother

\author{%
	Nicki S. Detlefsen \thanks{Section for Cognitive Systems, Technical University of Denmark} \\
	\texttt{nsde@dtu.dk}
	\And
	Søren Hauberg \printfnsymbol{1} \\
	\texttt{sohau@dtu.dk}
}

\begin{document}

\maketitle

%%%%%%%%%%% ABSTRACT %%%%%%%%%%%
\begin{abstract}
Disentangled representation learning finds compact, independent and easy-to-interpret factors of the data.
Learning such has been shown to require an inductive bias, which we explicitly encode in
a generative model of images. Specifically, we propose a model with two latent spaces:
one that represents spatial transformations of the input data, and another that represents
the transformed data. We find that the latter naturally captures the intrinsic appearance of the data. To realize the generative model, we propose a Variationally Inferred
Transformational Autoencoder (VITAE) that incorporates a spatial transformer into
a variational autoencoder.  We show how to perform inference in the model efficiently
by carefully designing the encoders and restricting the transformation class to be diffeomorphic. Empirically, our model separates the visual style from digit type on MNIST, separates shape and pose in images of human bodies and facial features from facial shape on CelebA. 

%Empirically, we find that our model outperform several general purpose disentanglement models on several image data sets.

\end{abstract}
%%%%%%%%%%% BODY %%%%%%%%%%%
\section{Introduction}

\emph{Disentangled Representation Learning (DRL)} is a fundamental challenge in machine learning that is currently seeing a renaissance within deep generative models. DRL approaches assume that an AI agent can benefit from separating out (disentangle) the underlying structure of data into disjointed parts of its representation. This can furthermore help interpretability of the decisions of the AI agent and thereby make them more accountable.

Even though there have been attempts to find a single formalized notion of disentanglement \cite{towards_a_definition_of_disentangled_representation}, no such theory exists (yet) which is widely accepted. However, the intuition is that a disentangled representation $\z$ should separate different informative factors of variation in the data \cite{representation_learning}. This means that changing a single latent dimension $z_i$ should only change a single interpretable feature in the data space $\Xcal$.

Within the DRL literature, there are two main approaches. The first is to hard-wire disentanglement into the model, thereby creating an inductive bias. This is well known \eg in convolutional neural networks, where the convolution operator creates an inductive bias towards translation in data. The second approach is to instead learn a representation
that is faithful to the underlying data structure, hoping that this is sufficient to disentangle the representation. However, there is currently little to no agreement in the literature on how to learn such representations \cite{challenging_common_assumptions}.

\begin{figure}
\centering
\begin{minipage}{0.65\textwidth}
  \centering
  \vspace{0.20cm}
  \includegraphics[width=1\textwidth, trim=2.85in 2.7in 4.2in 2.0in, clip]{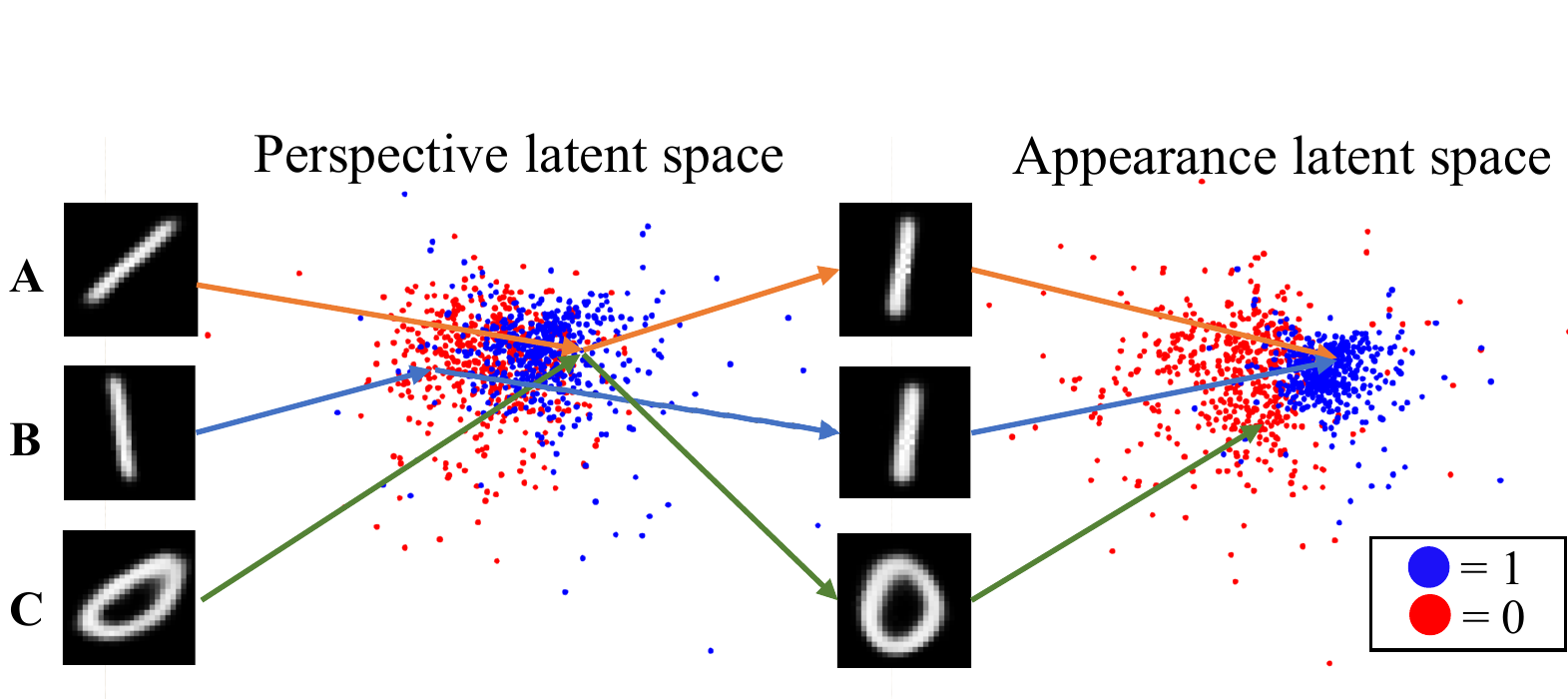}
  \vspace{0.25cm}
  \captionof{figure}{We disentangle data into \emph{appearance} and \emph{perspective} factors. First, data are encoded based on their \emph{perspective} (in this case image A and C are rotated in the same way), which is then removed from the original input. Hereafter, the transformed samples can be encoded in the \emph{appearance} space (image A and B are both ones), that encodes the factors left in data.}
  \label{fig:disentangle_explain}
\end{minipage} \hspace{0.5cm}
\begin{minipage}{0.3\textwidth}
  \centering
  \includegraphics[width=1\textwidth, trim=8.4cm 2.5cm 10cm 0cm, clip]{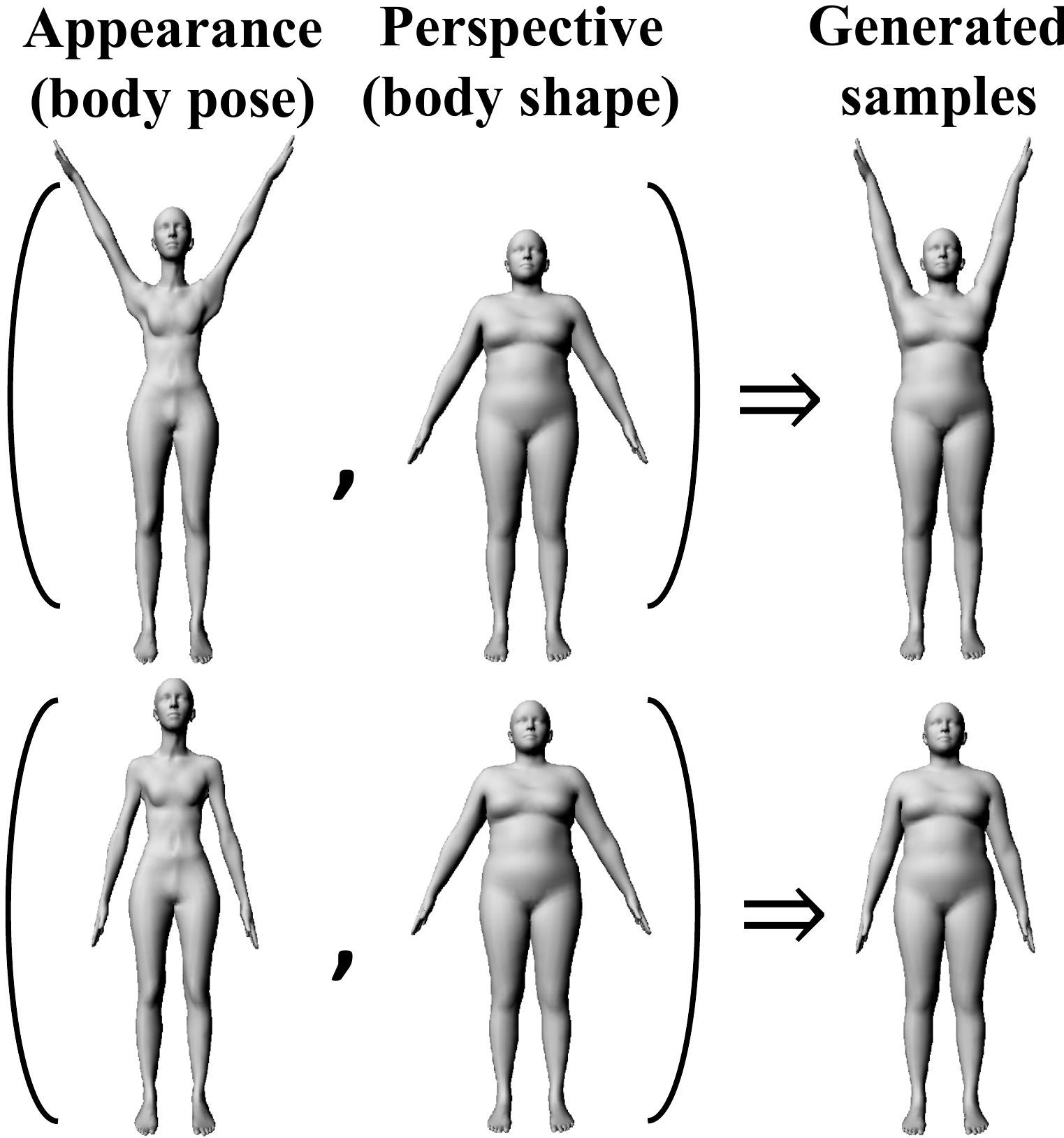}
  \captionof{figure}{Our model, VITAE, disentangles appearance from perspective. Here we separate body pose (arm position) from body shape.}
  \label{fig:front_fig}
\end{minipage}
\end{figure}

We consider disentanglement of two explicit groups of factors, the \emph{appearance} and the \emph{perspective}. We here define the appearance as being the factors of data that are left after transforming $\x$ by its perspective. Thus, the appearance is the \emph{form} or \emph{archetype} of an object and the perspective represents the specific realization of that archetype. Practically speaking, the perspective could correspond to an image rotation that is deemed irrelevant, while the appearance is a representation of the rotated image, which is then invariant to the perspective. This interpretation of the world goes back to Plato's allegory of the cave, from which we also borrow our terminology. This notion of removing perspective before looking at the appearance is well-studied within supervised learning, \eg using \emph{spatial transformer nets (STNs)} \cite{spatial_transformer_nets}.

\textbf{This paper contributes} an explicit model for disentanglement of appearance and perspective in images, called the \emph{variational inferred transformational autoencoder (VITAE)}. As the name suggests, we focus on variational autoencoders as generative models, but the idea is general (\FIG\ref{fig:disentangle_explain}). First we encode/decode the perspective features in order to extract an appearance that is perspective-invariant. This is then encoded into a second
latent space, where inputs with similar appearance are encoded similarly. This process generates an inductive bias that disentangles perspective and appearance. In practice, we develop an architecture that leverages the inference part of the model to guide the generator towards better disentanglement. We also show that this specific choice of architecture improves training stability with the right choice of parametrization of perspective factors. Experimentally, we demonstrate that our model on four datasets: standard disentanglement benchmark dSprites, disentanglement of style and content on MNIST, pose and shape on images of human bodies (\FIG\ref{fig:front_fig}) and facial features and facial shape on CelebA. 

% disentangle perspective from appearance content on MNIST and pose from shape on images of human bodies (\FIG\ref{fig:front_fig}).

\section{Related work}

\textbf{Disentangled representations learning (DRL)} have long been a goal in data analysis.
Early work on \emph{non-negative matrix factorization} \cite{lee1999learning} and \emph{bilinear models} \cite{Tenenbaum:2000} showed how images can be composed into semantic ``parts'' that can be glued together to form the final image. Similarly, \emph{EigenFaces} \cite{turk1991eigenfaces} have often been used to factor out lighting conditions from the representation \cite{shakunaga2001decomposed}, thereby discovering some of the physics that
govern the world of which the data is a glimpse.
This is central in the long-standing argument that for an AI agent to
understand and reason about the world, it must disentangle the explanatory factors
of variation in data \cite{building_machines_that_learn_and_think_like_people}.
As such, DRL can be seen as a poor man's approximation
to discovering the underlying causal factors of the data.

\textbf{Independent components} are, perhaps, the
most stringent formalization of ``disentanglement''. The seminal
\emph{independent component analysis (ICA)} \cite{independent_component_analysis}
factors the signal into statistically independent components.
It has been shown that the independent components of \emph{natural images} are edge
filters \cite{bell1997independent} that can be linked to the receptive
fields in the human brain \cite{olshausen1996emergence}.
Similar findings have been made for both \emph{video} and \emph{audio}
\cite{van1998independent, lewicki2002efficient}.
DRL, thus, allows us to understand both the data and ourselves.
Since independent factors are the optimal compression, ICA finds the most compact
representation, implying that the predictive model can achieve maximal capacity
from its parameters.
This gives DLR a predictive perspective, and can be taken as a hint that a well-trained
model might be disentangled.
In the linear case, independent components have many successful realizations
\cite{hyvarinen2000independent}, but in the general non-linear case, the problem
is not identifiable \cite{nonlinear_ICA}.
%Related, \citet{challenging_common_assumptions} have shown that DRL is generally not possible without an inductive bias in both the model and the data.

\textbf{Deep DRL}
was initiated by \citet{representation_learning} who sparked the current interest in the topic.
One of the current state-of-the-art methods for doing disentangled representation
learning is the $\beta$-VAE \citep{beta_vae}, that modifies the
\emph{variational autoencoder (VAE)} \cite{auto_encoding_variational_bayes, stochastic_backpropagation_and_approximate_inference}
to learn a more disentangled representation. $\beta$-VAE enforces more weight on the KL-divergence in the VAE loss, thereby
optimizing towards latent factors that should be axis aligned \ie disentangled.
%However, \citet{challenging_common_assumptions} have recently show that it is fundamentally impossible to do DLR without some kind of inductive bias.
%\citet{beta_vae} also propose a measure of disentanglement, but this has been
%shown to neither be general nor unbiased \cite{isolating_sources_of_disentanglement}. 
Newer models like $\beta$-TCVAE \citep{Chen2018} and DIP-VAE \citep{Kumar2017}
extend $\beta$-VAE by decomposing the KL-divergences into multiple terms, and only
increase the weight on terms that analytically disentangles the models.
% Other popular deep models worth mentioning that still performs disentanglement learning, but not by adjusting the KL-term, are the InfoGAN model by \citet{infogan} and the DC-IGN model by \citet{dc_ign}. 
InfoGAN \citep{infogan} extends the latent code $\z$ of the standard GAN model
\cite{generative_adversarial_nets} with an extra latent code $c$ and then penalize
low  mutual information between generated samples $G(c, z)$ and $c$.
DC-IGN \citep{dc_ign} forces the latent codes to be disentangled by only feeding
in batches of data that vary in one way (\eg pose, light) while only having small
disjoint parts of the latent code active. % at the same time. 

\textbf{Shape statistics} is the key inspiration for our work.
The shape of an object was first formalized by
\citet{kendall1989survey} as being what is left of an object
when \emph{translation}, \emph{rotation} and \emph{scale} are factored out.
That is, the intrinsic shape of an object should not depend on viewpoint.
This idea dates, at least, back to \citet{thompson1942growth}
who pioneered the understanding of the development of biological forms.
In Kendall's formalism, the rigid transformations ({translation}, {rotation} and {scale})
are viewed as group actions to be factored out of the representation, such that
the remainder is \emph{shape}.
\citet{towards_a_definition_of_disentangled_representation} follow the same idea by defining
disentanglement as a factoring of the representation into group actions. Our work can 
be seen as a realization of this principle within a deep generative model. When an object is represented by a set of landmarks, \eg in the form of discrete
points along its contour, then Kendall's \emph{shape space} is a Riemannian
manifold that exactly captures all variability among the landmarks except
translation, rotation, and scale of the object. When the object is not represented
by landmarks, then similar mathematical results are not available.
Our work shows how the same idea can be realized for general image data, and for a much
wider range of transformations than the rigid ones.
\citet{miller2006learned} proposed a related linear model that generate new data by 
transforming a prototype, which is estimated by joint alignment.

\textbf{Transformations} are at the core of our method, and these leverage the architecture
of spatial transformer nets (STNs) \cite{spatial_transformer_nets}. While these
work well within supervised learning, \cite{inverse_compositional_spatial_transformer_networks,
densely_fused_spatial_transformer_networks, deep_diffeomorphic_transformer_networks}
there has been limited uptake within generative models. \citet{ST_GAN} combine a GAN %\cite{generative_adversarial_nets} 
with an STN to compose a foreground (e.g a furniture) into a background such
that it look neutral.
The AIR model \citep{Eslami2016} combines STNs with a VAE for object rendering,
but do not seek disentangled representations. In supervised learning, \emph{data augmentation} is often used to make a classifier partially invariant to select transformations \citep{baird1992document, hauberg:aistats:2016}.

\section{Method}
%Our model's main components are variational autoencoders and spatial transformer nets, which we now describe.

\begin{figure}
\centering
\subfloat[VAE]{\includegraphics[width=0.30\textwidth, clip]{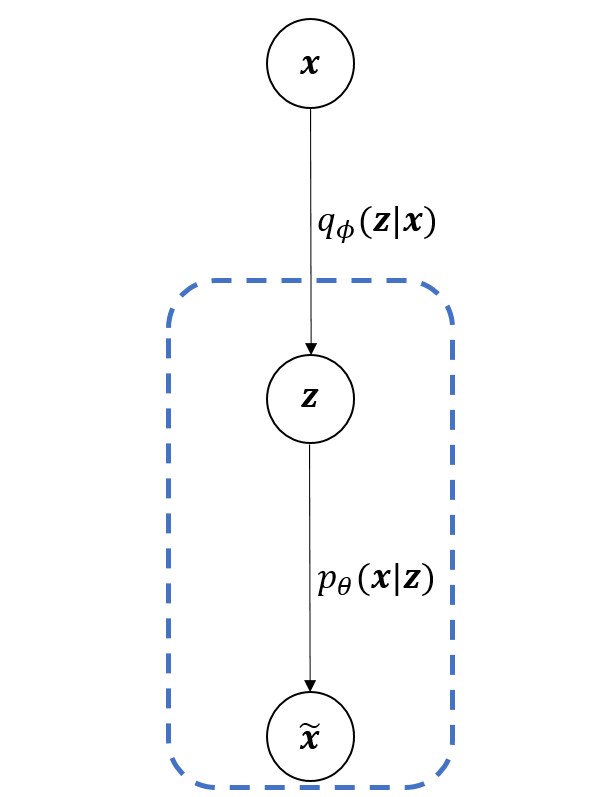} \label{fig:vae}} \hspace{-0.5cm}
\subfloat[Unconditional VITAE]{\includegraphics[width=0.30\textwidth]{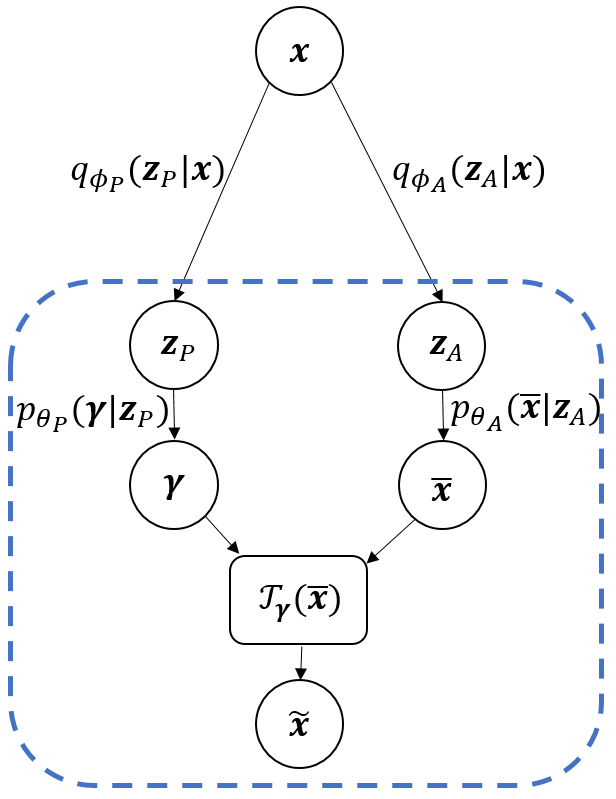} \label{fig:vitae_ui}} \hspace{0.5cm}
\subfloat[Conditional VITAE]{\includegraphics[width=0.30\textwidth]{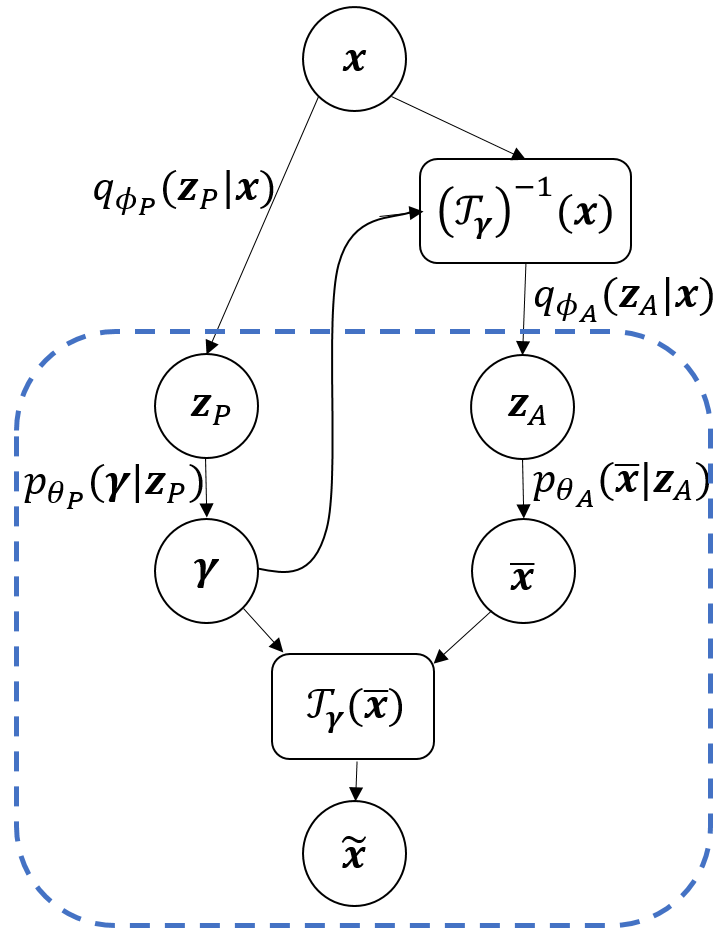} \label{fig:vitae_ci}}
\caption{Architectures of standard VAE and our proposed U-VITAE and C-VITAE models. Here $q$ denotes encoders, $p$ denotes decoders, $\Tcal^\bgamma$ denotes a ST-layer with transformation parameters $\bgamma$. The dotted box indicates the generative model. }
\label{fig:architechtures}
\end{figure}

Our goal is to extend a variational autoencoder (VAE) \cite{auto_encoding_variational_bayes, stochastic_backpropagation_and_approximate_inference} such that it can disentangle appearance and perspective in data. A standard VAE  assumes that data is generated by a set of latent variables following a standard Gaussian prior,
\begin{align}
\begin{split}
p(\x) &= \int p(\x | \z) p(\z) \mathrm{d}\z \\
p(\z) &= \Ncal(\textbf{0}, \mathbb{I}_d), \: p(\x | \z) = \Ncal(\x | \bmu_p(\z) , \bsigma_p^2 (\z)) \: \text{or}\:  P(\x | \z) = \mathcal{B}(\x | \bmu_p(\z)).
\end{split}
\end{align}
Data $\x$ is then generated by first sampling a latent variable $\z$ and then
sample $\x$ from the conditional $p(\x | \z)$ (often called the decoder).
To make the model flexible enough to capture complex data distributions,
$\bmu_p$ and $\bsigma_p^2$ are modeled as deep neural nets. The marginal likelihood
is then intractable and a variational approximation $q$ to $p(\z | \x)$ is needed,
\begin{equation}
p(\z | \x) \approx q(\z | \x) = \Ncal(\z | \bmu_q(\x) , \bsigma_q^2 (\x)),
\end{equation}
where $\bmu_q(\x)$ and $\bsigma_q^2(\x)$ are deep neural networks, see \FIG\ref{fig:vae}.

When training VAEs, we therefore simultaneously train a generative model $p_\theta (\x | \z) p_\theta (\z)$ and an inference model $q_\phi(\z | \x)$ (often called the encoder). This is done by maximizing a variational lower bound to the likelihood $p(\x)$ called the \emph{evidence lower bound (ELBO)}
\begin{align}
\label{eq:elbo} \log p(\x) &\geq \mathbb{E}_{q_\phi(\z | \x)} \left[ \log \dfrac{p_\theta(\x, \z)}{q_\phi(\z | \x)}  \right] 
=\underbrace{\mathbb{E}_{q_\phi(\z | \x)} \left[ \log p_\theta (\x | \z) \right]}_{\text{data fitting term}}-\underbrace{KL(q_\phi(\z | \x) || p_\theta(\z))}_{\text{regulazation term}}. 
\end{align}
The first term measures the reconstruction error between $\x$ and $p_\theta(\x | \z)$
and the second measures the KL-divergence between the encoder $q_\phi(\z | \x)$ and
the prior $p(\z)$. \EQN\ref{eq:elbo} can be optimized using the reparametrization trick
\cite{auto_encoding_variational_bayes}. Several improvements to VAEs have been
proposed \cite{importance_weighted_autoencoders, improving_variational_inference_with_inverse_autoregressive_flow},
but our focus is on the standard model.

\subsection{Incorporating an inductive bias}
To incorporate an inductive bias that is able to disentangle appearance from perspective, we change the underlying generative model to rely on two latent factors $\z_A$ and $\z_P$,
\begin{equation}
\label{eq:genrative_model}
p(\x) = \iint p(\x | \z_A, \z_P ) p(\z_A) p(\z_P) \mathrm{d}\z_A \mathrm{d}\z_P,
\end{equation}
where we assume that $\z_A$ and $\z_P$ both follow standard Gaussian priors.
Similar to a VAE, we also model the generators as deep neural networks. To generate new data $\x$, we combine the appearance and perspective factors using the following 3-step procedure that uses a spatial transformer (ST) layer \cite{spatial_transformer_nets} (dotted box in \FIG\ref{fig:vitae_ui}):

\begin{enumerate}
\item Sample $\z_A$ and $\z_P$ from $p(\z) = \Ncal(\textbf{0}, \mathbb{I}_d)$.
\item Decode both samples $\tilde{\x} \sim p(\x | \z_A)$, $\bgamma \sim p(\x | \z_P)$. 
\item Transform $\tilde{\x}$ with parameters $\bgamma$ using a spatial transformer layer: $\x = \Tcal_\gamma(\tilde{\x})$. 
\end{enumerate}
This process is illustrated by the dotted box in \FIG\ref{fig:vitae_ui}. 

\paragraph{Unconditional VITAE inference.}
As the marginal likelihood \eqref{eq:genrative_model} is intractable, we use variational inference. A natural choice is to approximate each latent group of factors $\z_A, \z_P$ independently of the other \ie
\begin{equation}
\label{eq:vitae_ui} p(\z_P | \x) \approx q_P(\z_P | \x) \; \text{and} \; p(\z_A | \x) \approx q_A(\z_A | \x).
\end{equation}
The combined inference and generative model is illustrated in \FIG\ref{fig:vitae_ui}. For comparison, a VAE model is shown in \FIG\ref{fig:vae}. It can easily be shown that the ELBO for this model is merely a VAE with a KL-term for each latent space (see supplements).
%\begin{equation}
%\label{eq:elbo_vitae_ui}
%\log p(\x) \geq E_{q_A, q_P} [\log(p(\x | \z_A, \z_P)] -D_{KL}(q_P(\z_P | \x) || p(\z_P)) -D_{KL}(q_A(\z_A | \x) || p(\z_A)),
%\end{equation}
%which is merely a VAE with a KL-term for each latent space. The derivation can be found in \SUPMAT.

\paragraph{Conditional VITAE inference.}
This inference model does \emph{not} mimic the generative process of the model,
which may be suboptimal. Intuitively, we expect the encoder to approximately 
perform the inverse operation of the decoder, \ie
$\z \approx \text{encoder}(\text{decoder}(\z)) \approx \text{decoder}^{-1}(\text{decoder}(\z))$.
Since the proposed encoder \eqref{eq:vitae_ui} does not include an ST-layer, it may
be difficult to train an encoder to approximately invert the decoder.
To accommodate this, we first include an ST-layer in the encoder for the appearance
factors. Secondly, we explicitly enforce that the predicted transformation in the
encoder $\Tcal^{\bgamma_e}$ is the inverse of that of the decoder $\Tcal^{\bgamma_d}$,
\ie $\Tcal^{\bgamma_e}=(\Tcal^{\bgamma_d})^{-1}$ (more on invertibility  
in Sec.~\ref{sec:transformer_class}).
The inference of appearance is now dependent on the perspective factor $\z_P$, \ie
\begin{align}
  \label{eq:vitae_ci}
   p(\z_P | \x) \approx q_P(\z_P | \x) \; \text{and} \; p(\z_A | \x) \approx q_A(\z_A | \x, \z_P) .
\end{align}
These changes to the inference architecture are illustrated in \FIG\ref{fig:vitae_ci}. It can easily be shown that the ELBO for this model is given by
\begin{equation}
\label{eq:elbo_vitae_ci}
\log p(\x) \geq \mathbb{E}_{q_A, q_P} [\log(p(\x | \z_A, \z_P)] -D_{KL}(q_P(\z_P | \x) || p(\z_P)) - \mathbb{E}_{q_P} [D_{KL}(q_A(\z_A | \x) || p(\z_A))] .
\end{equation}
which resembles the standard ELBO with a additional term (derivation in \SUPMAT), corresponding to the second latent space. We will call both models \textit{variational inferred transformational autoencoders (VITAE)} and we will denote the first model \eqref{eq:vitae_ui} as \emph{unconditional/U-VITAE} and the second model \eqref{eq:vitae_ci} as \emph{conditional/C-VITAE}. The naming comes from \EQN\ref{eq:vitae_ui} and \ref{eq:vitae_ci}, where $\z_A$ is respectively unconditioned and conditioned on $\z_P$. Experiments will show that the conditional architecture is essential for inference (Sec.~\ref{sec:mnist_experiments}).

\subsection{Transformation classes}
\label{sec:transformer_class}

Until now, we have assumed that there exists a class of transformations $\Tcal$
that captures the perspective factors in data. Clearly, the choice of $\Tcal$ depends on
the true factors underlying the data, but in many cases an affine transformation should suffice.
\begin{wrapfigure}[23]{r}{0.4\textwidth}
	\centering
	\subfloat{\includegraphics[width=0.3\textwidth,trim={1in 1in 1in 1in},clip]{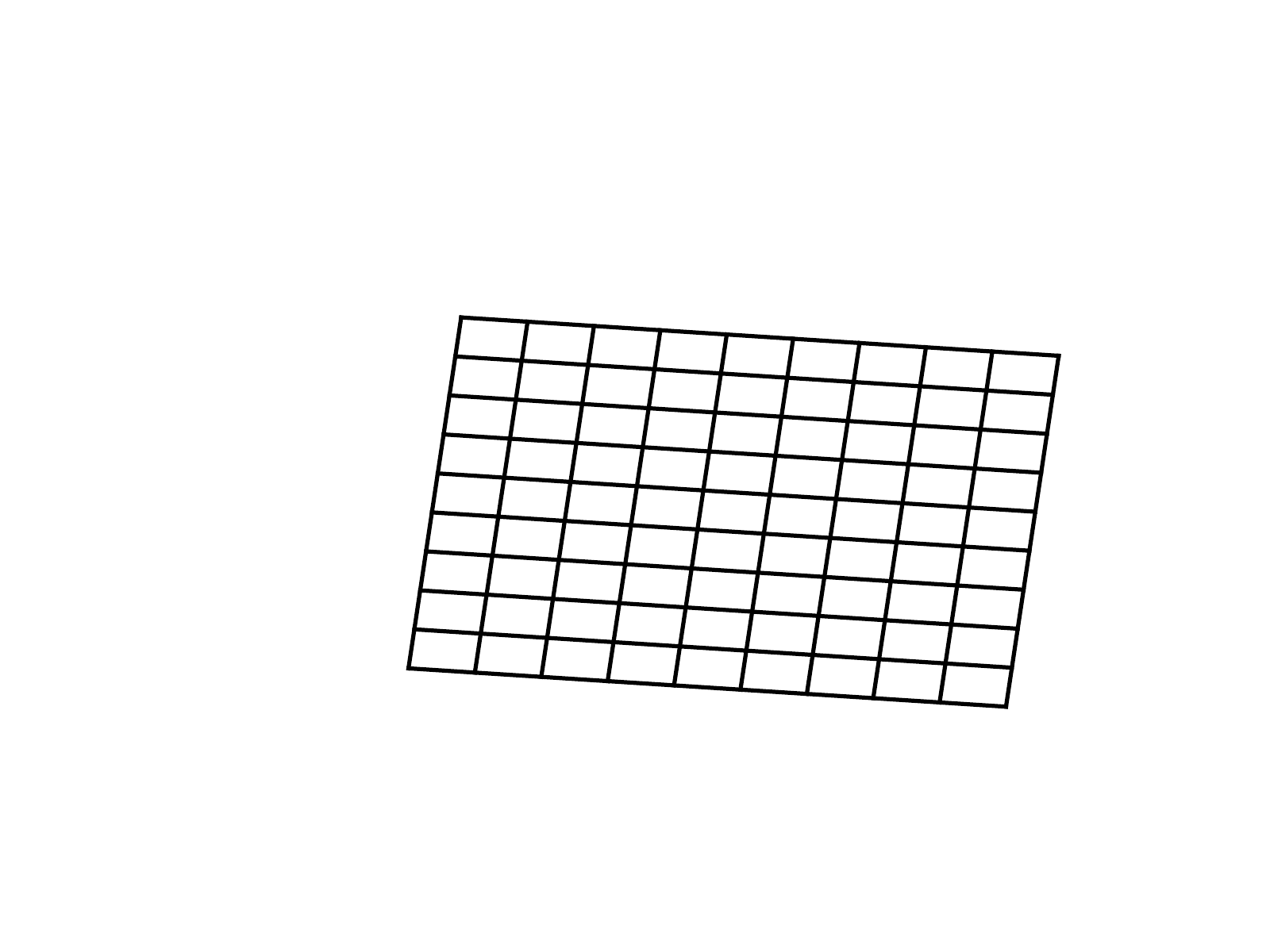}} \hspace{-1cm}
	\subfloat{\includegraphics[width=0.3\textwidth,trim={1in 1in 1in 1in},clip]{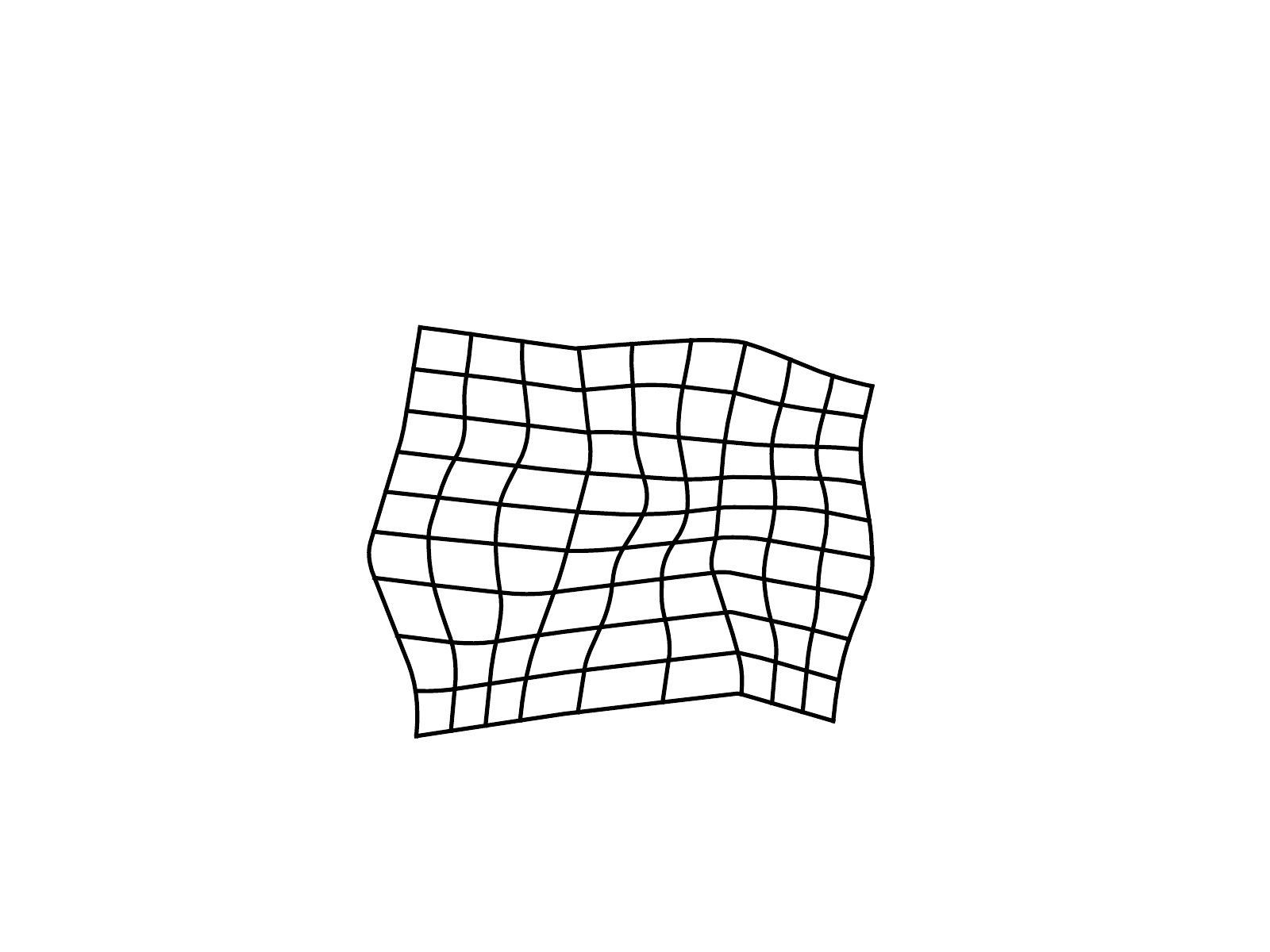}}
  \caption{Random deformation field of an affine transformation (top) compared to a CPAB (bottom). We clearly see that CPAB transformations offers a mush more flexible and rich class of diffiomorphic transformations.}
\vspace{-1.5cm}
\label{fig:vec_compare}
\end{wrapfigure}
\begin{equation}
\label{eq:affine}
\Tcal_\bgamma (\x) = \mat{A} \x + \vec{b} = \begin{bmatrix}
\gamma_{11} & \gamma_{12} & \gamma_{13} \\
\gamma_{21} & \gamma_{22} & \gamma_{14}
\end{bmatrix} \begin{bmatrix}
x \\ y \\ 1
\end{bmatrix}.
\end{equation}
However, the C-VITAE model requires access to the inverse 
transformation $ \Tcal^{-1}$. The inverse of \EQN \ref{eq:affine} is given by
$\Tcal_\bgamma^{-1} (\x) = \mat{A}^{-1}\x - \vec{b}$, which only exist if $\mat{A}$ has a non-zero determinant.

One, easily verified, approach  to secure invertibility is to parametrize the transformation by
two scale factors $s_x, s_y$, one rotation angle $\alpha$, one shear parameter $m$
and two translation parameters $t_x, t_y$:
\begin{equation}
\label{eq:affinedecomp}
\Tcal_\bgamma(\x) = \begin{bmatrix}
\cos(\alpha) & -\sin(\alpha) \\
\sin(\alpha) & \;\;\;\cos(\alpha) 
\end{bmatrix}
\begin{bmatrix}
1 & m\\
0 & 1
\end{bmatrix} \begin{bmatrix}
s_x & 0 \\
0 & s_y
\end{bmatrix} + \begin{bmatrix}
t_x \\ t_y
\end{bmatrix}.
\end{equation}
In this case the inverse is trivially
\begin{equation}
\Tcal_{(s_x, s_y, \gamma, m, t_x, t_y)}^{-1} (\x)= \Tcal_{(\frac{1}{s_x}, \frac{1}{s_y}, -\gamma, -m, -t_x, -t_y)} (\x),
\end{equation}
where the scale factors must be strictly positive.

An easier and more elegant approach is to leverage the matrix exponential.
That is, instead of parametrizing the transformation in \EQN\ref{eq:affine}, we
instead parametrize the velocity of the transformation
\begin{equation}
\label{eq:affinediff}
\Tcal_\bgamma (\x) = \textbf{expm} \left(\begin{bmatrix}
\gamma_{11} & \gamma_{12} & \gamma_{13} \\
\gamma_{21} & \gamma_{22} & \gamma_{14} \\
0 & 0 & 0
\end{bmatrix} \right) \begin{bmatrix}
x \\ y \\ 1
\end{bmatrix}.
\end{equation}
The inverse\footnote{Follows from $\Tcal_{\bgamma}$ and $\Tcal_{-\bgamma}$ being commuting matrices.}
is then $\Tcal_{\bgamma}^{-1} = \Tcal_{-\bgamma}$. Then $\Tcal$ in \EQN\ref{eq:affinediff} is a $C^\infty$-diffiomorphism
(\ie a differentiable invertible map with a differentiable inverse) \citep{lie_algebra}. Experiments show that diffeomorphic transformations stabilize training and yield tighter ELBOs (see supplements).

%The fact that $\Tcal_{\gamma}$ becomes a diffiomorphism also has positive optimization properties. \FIG\ref{fig:stability} shows the ELBO as a function of the learning rate $\lambda$ for the three different choices of affine parametrization, using our C-VITAE architecture. We clearly see that the diffeomorphic affine parametrization archives a tighter bound, and can run for much higher learning rates (faster convergence) before the network begins to diverge. These findings are similar to those of \citet{deep_diffeomorphic_transformer_networks} in the supervised context.

Often we will not have prior knowledge regarding which transformation classes
are suitable for disentangling the data. A natural way forward is then to apply
a highly flexible class of transformations that are treated as ``black-box''.
Inspired by \citet{deep_diffeomorphic_transformer_networks}, we also consider
transformations $\Tcal_{\bgamma}$ using the highly expressive diffiomorphic
transformations \textit{CPAB} from \citet{highly_expressive_spaces_of_well_behaved_transformations}. 
These can be viewed as an extension to \EQN\ref{eq:affinediff}:
instead of having a single affine transformation parametrized by its velocity,
the image domain is divided into smaller cells, each having their own affine velocity.
The collection of local affine velocities can be efficiently parametrized and integrated,
giving a fast and flexible diffeomorphic transformation, see \FIG\ref{fig:vec_compare} for a comparison between an affine transformation and a CPAB transformation. For details, see \citet{highly_expressive_spaces_of_well_behaved_transformations}.

We note, that our transformer architecture are similar to the work of \citet{lorenz19} and \citet{xing19} in that they also tries to achieve disentanglement through spatial transformations. However, our work differ in the choice of transformation. This is key, as the theory of \citet{towards_a_definition_of_disentangled_representation} strongly relies on disentanglement through \textit{group actions}. This places hard constrains on which spatial transformations are allowed: \textit{they have to form a smooth group}. Both thin-plate-spline transformations considered in \citet{lorenz19} and displacement fields considered in \citet{xing19} are not invertible and hence do not correspond to proper group actions. Since diffiomorphic transformations form a smooth group, this choice is paramount to realize the theory of \citet{towards_a_definition_of_disentangled_representation}.
%
%As outlined earlier, we only consider diffiomorphic transformation that indeed form such smooth group, and our work therefore remains unique since it is the first practical realization of the theory of \citet{towards_a_definition_of_disentangled_representation}.
\section{Experimental results and discussion}

For all experiments, we train a standard VAE, a $\beta$-VAE \citep{beta_vae}, a $\beta$-TCVAE \citep{Chen2018}, a DIP-VAE-II \citep{Kumar2017} and our developed VITAE model. We model the encoders and decoders as multilayer perceptron networks (MLPs). For a fair comparison, the number of trainable parameters is approximately the same in all models. The models were implemented in Pytorch \cite{pytorch} and the code is available at \url{https://github.com/SkafteNicki/unsuper/}.

\textbf{Evaluation metric.} Measuring disentanglement still seems to be an unsolved problem, but the work of \citet{challenging_common_assumptions} found that most proposed disentanglement metrics are highly correlated. We have chosen to focus on the DIC-metric from \citet{Eastwood2019}, since this metric has seen some uptake in the research community. This metric measures how will the generative factors can be predicted from latent factors. For the MNIST and SMPL datasets, the generative factors are discrete instead of continuous, so we change the standard linear regression network to a kNN-classification algorithm. We denote this metric $D_{score}$ in the results.

\begin{figure}[b]
\centering
\subfloat{
	\includegraphics[width=0.35\textwidth, height=1cm]{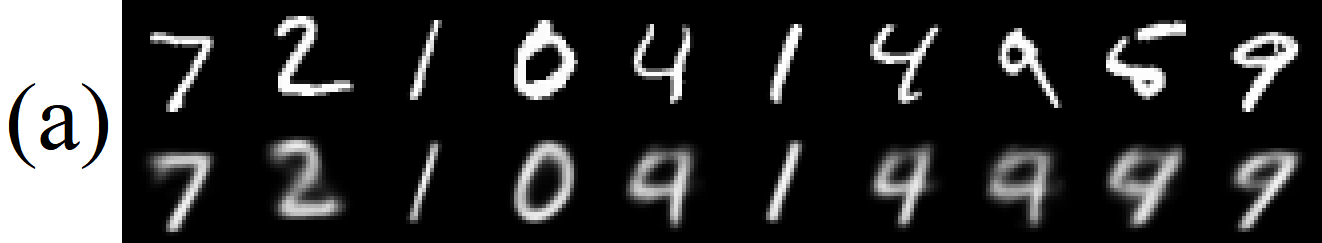}
	\includegraphics[width=0.65\textwidth, height=1cm]{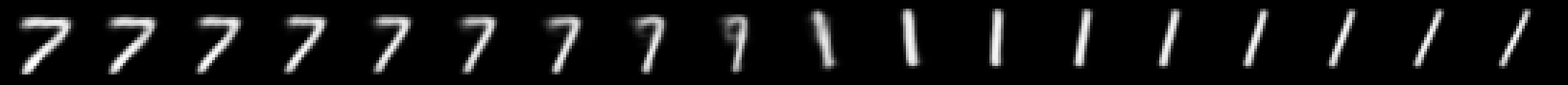}
	\label{fig:vae_mani}
} \\ \vspace{-0.25cm}
\subfloat{
	\includegraphics[width=0.35\textwidth, height=1cm]{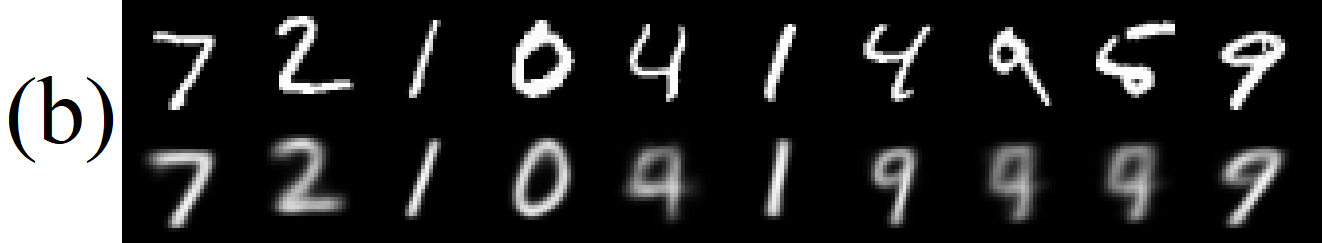}
	\includegraphics[width=0.65\textwidth, height=1cm]{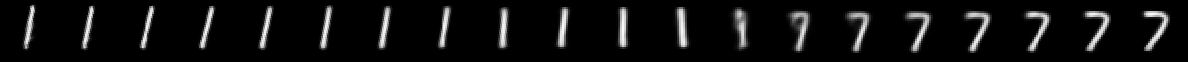}
	\label{fig:betavae_mani}
} \\ \vspace{-0.25cm}
\subfloat{
	\includegraphics[width=0.35\textwidth, height=1cm]{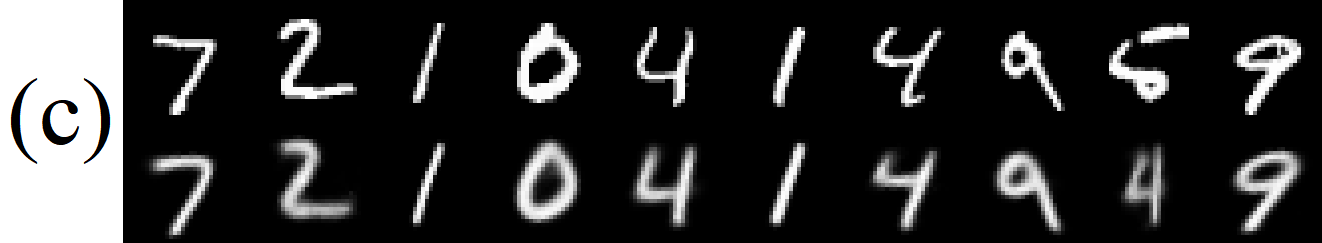}
	\includegraphics[width=0.65\textwidth, height=1cm]{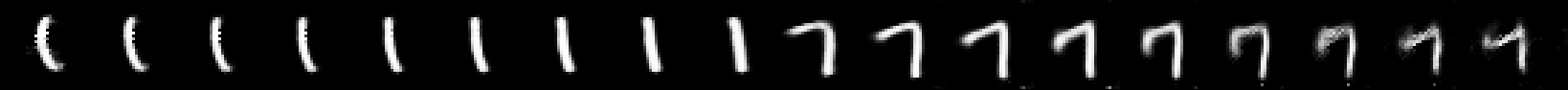}
	\label{fig:vitae_mani}
} \\ \vspace{-0.38cm}
\subfloat{
	\includegraphics[width=0.65\textwidth, height=1cm]{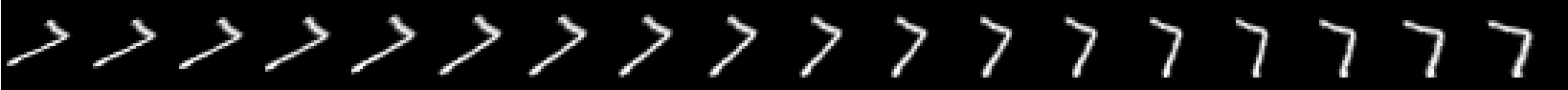}
} \hspace{-5.222cm}
\caption{Reconstructions (left images) and manipulation of latent codes (right images) on MNIST for the three different models: VAE (a), $\beta$-VAE (b) and C-VITAE (c). The right images are generated by varying one latent dimension in all models, while keeping the rest fixed. For the C-VITAE model, we have shown this for both the appearance and perspective spaces.}\label{fig:interpolation_res}
\end{figure}

\begin{table}[]
\resizebox{\textwidth}{!}{
\setlength\tabcolsep{3pt}
\begin{tabular}{l|lll|lll|lll|}
              & \multicolumn{3}{c|}{dSprite}                                                                   & \multicolumn{3}{c|}{MNIST}                                                                     & \multicolumn{3}{c|}{SMPL}                                                                             \\ \hline
              & \multicolumn{1}{c}{ELBO} & \multicolumn{1}{c}{$\log p(\x)$} & \multicolumn{1}{c|}{$D_{score}$} & \multicolumn{1}{c}{ELBO} & \multicolumn{1}{c}{$\log p(\x)$} & \multicolumn{1}{c|}{$D_{score}$} & \multicolumn{1}{c}{ELBO}        & \multicolumn{1}{c}{$\log p(\x)$} & \multicolumn{1}{c|}{$D_{score}$} \\
VAE           & -47.05                   & -49.32                           & 0.05                             & -169                     & -172                             & 0.579                            & $-8.62\times 10^3$              & $-8.62\times 10^3$               & 0.485                            \\
$\beta$-VAE   & -79.45                   & -81.38                           & 0.18                             & -150                     & -152                             & 0.653                            & $-8.62\times 10^3$              & $-8.60\times 10^3$               & 0.525                            \\
$\beta$-TCVAE & -66.48                   & -68.12                           & 0.30                             & -141                     & -144                             & 0.679                            & $-8.62\times 10^3$              & $-8.56\times 10^3 $              & 0.651                            \\
DIP-VAE-II    & $\textbf{-46.32}$        & $\textbf{-48.92}$                & 0.12                             & -140                     & -155                             & 0.733                            & $-8.62\times 10^3$              & $-8.54\times 10^3 $              & 0.743                            \\
U-VITAE       & -55.25                   & -57.29                           & 0.22                             & -142                     & -143                             & 0.782                            & $-8.62\times 10^3$              & $-8.55\times 10^3 $              & 0.673                            \\
C-VITAE       & -68.26                   & -70.49                           & $\textbf{0.38}$                  & $\textbf{-139}$          & $\textbf{-141}$                  & $\textbf{0.884}$                 & $\boldsymbol{-8.62\times 10^3}$ & $\boldsymbol{-8.52\times 10^3}$  & $\textbf{0.943}$                
\end{tabular}
}
\vspace{0.1cm}
\caption{Quantitative results on three datasets. For each dataset we report the ELBO, test set log likelihood and disentanglement score $D_{score}$. Bold marks best results.}
\label{tab:all_quantitative_res}
\end{table}

\subsection{Disentanglement on shapes}
\label{sec:dsprite_experiments}
We initially test our models on the dSprites dataset \citep{dsprites17}, which is a well established disentanglement benchmarking dataset to evaluate the performance of disentanglement algorithms. The results can be seen in Table \ref{tab:all_quantitative_res}. We find that our proposed C-VITAE model perform best, followed by the $\beta$-TCVAE model in terms of disentanglement. The experiments clearly shows the effect on performance of the improved inference structure of C-VITAE compared to U-VITAE. It can be shown that the conditional architecture of C-VITAE, minimizes the mutual information between $\z_A$ and $\z_P$, leading to better disentanglement of the two latent spaces. To get the U-VITAE architecture to work similarly would require a auxiliary loss term added to the ELBO. 

\subsection{Disentanglement of MNIST images}
\label{sec:mnist_experiments}

Secondly, we test our model on the MNIST dataset \cite{MNIST}. To make the task more difficult, we artificially augment the dataset by first randomly rotating each image by an angle uniformly chosen in the interval $[-20^\circ , 20^\circ]$ and secondly translating the images by $t = [x,y]$, where $x,y$ is uniformly chosen from the interval [-3, 3]. For VITAE, we model the perspective with an affine diffiomorphic transformation (\EQN\ref{eq:affinediff}).  

The quantitative results can be seen in \TAB\ref{tab:all_quantitative_res}. We clearly see that C-VITAE outperforms the alternatives on all measures. We overall observes that better disentanglement, seems to give better distribution fitting. Qualitatively, \FIG \ref{fig:interpolation_res} shows the effect of manipulating the latent codes alongside test reconstructions for VAE, $\beta$-VAE and C-VITAE. Due to space constraints, the results from $\beta$-TCVAE and DIP-VAE-II can found in the supplementary material. The plots were generated by following the protocol from \citet{beta_vae}: one latent factor is linearly increased from -3 to 3, while the rest is kept fixed. In the VAE (\FIG\ref{fig:vae_mani}), this changes both the appearance (going from a 7 to a 1) and the perspective (going from rotated slightly left to rotated right).
We see no meaningful disentanglement of latent factors.
In the $\beta$-VAE model (\FIG\ref{fig:betavae_mani}), we observe some disentanglement,
since only the appearance changes with the latent factor. However this disentanglement
comes at the cost of poor reconstructions. This trade-off is directly linked to
the emphasized regularization in the $\beta$-VAE. We note that the value $\beta=4.0$
proposed in the original paper \cite{beta_vae} is insufficiently low for our
experiments to observe any disentanglement, and we use $\beta=8.0$ based on
qualitative evaluation of results. For $\beta$-TCVAE and DIP-VAE-II we observe nearly the same amount of qualitative disentanglement as $\beta$-VAE, however these models achieve less blurred samples and reconstructions. This is probably due to the two models decomposition of the KL-term, only increasing the parts that actually contributes to disentanglement. Finally, for our developed VITAE model (\FIG \ref{fig:vitae_mani}),
we clearly see that when we change the latent code in the appearance space (top row),
we only change the content of the generated images, while manipulating the
latent code in the perspective space (bottom row) only changes the perspective \ie image orientation.

Interestingly, we observe that there exists more than one prototype of a 1 in
the appearance space of VITAE, going from slightly bent to straightened out.
By our definition of disentanglement, that \emph{everything left} after
transforming the image is appearance, there is nothing wrong with this. This is
simply a consequence of using an affine transformation that cannot model this kind
of local deformation. Choosing a more flexible transformation class could factor
out this kind of perspective. The supplements contain generated samples from the different models.

\subsection{Disentanglement of body shape and pose}
\label{sec:body_experiments}

We now consider synthetic image data of human bodies generated by the
\emph{Skinned Multi-Person Linear Model (SMPL)} \cite{SMPL} which are explicitly
factored into \emph{shape} and \emph{pose}. We generate 10,000
bodies (8,000 for training, 2,000 for testing), by first continuously sampling
body shape (going from thin to thick) and then uniformly sampling a body pose from
four categories ((arms up, tight), (arms up, wide), (arms down, tight), (arms down, wide)).
\FIG\ref{fig:front_fig} shows examples of generated images. Since change in body
shape approximately amounts to a local shape deformation, we model the perspective
factors using the aforementioned "black-box" diffiomorphic CPAB transformations (Sec. \ref{sec:transformer_class}). The remaining appearance factor should then reflect body pose.

%\begin{wraptable}[10]{R}{7cm}
%\vspace{-4mm}
%\centering
%\begin{tabular}{lcc}
%\multicolumn{1}{l|}{\textbf{Model}} 	& $\log p(\x)$ 				& $D_{score}$	\\ \hline
%\multicolumn{1}{l|}{VAE}     		    & $-8.62\times 10^3$ 		& 0.485\\ %\hline
%\multicolumn{1}{l|}{$\beta$-VAE}    	& $-8.60\times 10^3$		& 0.525\\ %\hline
%\multicolumn{1}{l|}{$\beta$-TCVAE}		& $-8.56\times 10^3 $    	& 0.651\\
%\multicolumn{1}{l|}{DIP-VAE-II}			& $-8.54\times 10^3 $ 		& 0.743\\
%\multicolumn{1}{l|}{C-VITAE} 		    & $\boldsymbol{-8.52\times 10^3}$ & $\boldsymbol{0.943}$
%\end{tabular}
%\vspace{0.1cm}
%\caption{Test set log likelihood and disentanglement score for the different models on SMPL.}
%\label{tab:smpl_table}
%\end{wraptable}

\textbf{Quantitative evaluation.} We again refer to \TAB\ref{tab:all_quantitative_res} that shows ELBO, test set log-likelihood and disentanglement score for all models. 
%For this dataset, we again apply the disentanglement metric from \citet{Eastwood2019} to work discrete generative factors (body pose). We denote this $D_{score}$, and it measures how well the different models latent spaces can predict the true generative factors. In \TAB\ref{tab:smpl_table} show the test set log-likelihood and $D_{score}$ for all models. 
As before, C-VITAE is both better at modelling the data distribution and achieves a higher disentanglement score. The explanation is that for a standard VAE model (or $\beta$-VAE and its variants for that sake) to learn a complex body shape deformation model, it requires a high capacity network. However, the VITAE architecture gives the autoencoder a short-cut to learning these transformations that only requires optimizing a few parameters. We are not guaranteed that the model will learn anything meaningful or that it actually uses this short-cut, but experimental evidence points in that direction.
A similar argument holds in the case of MNIST, where a standard MLP may struggle to learn rotation of digits, but the ST-layer in the VITAE architecture provides a short-cut. Furthermore, we found the training of VITAE to be more stable than other models. 

\begin{figure}[h!]
\centering
\includegraphics[width=1\textwidth, trim=0cm 0cm 12cm 6.47cm, clip]{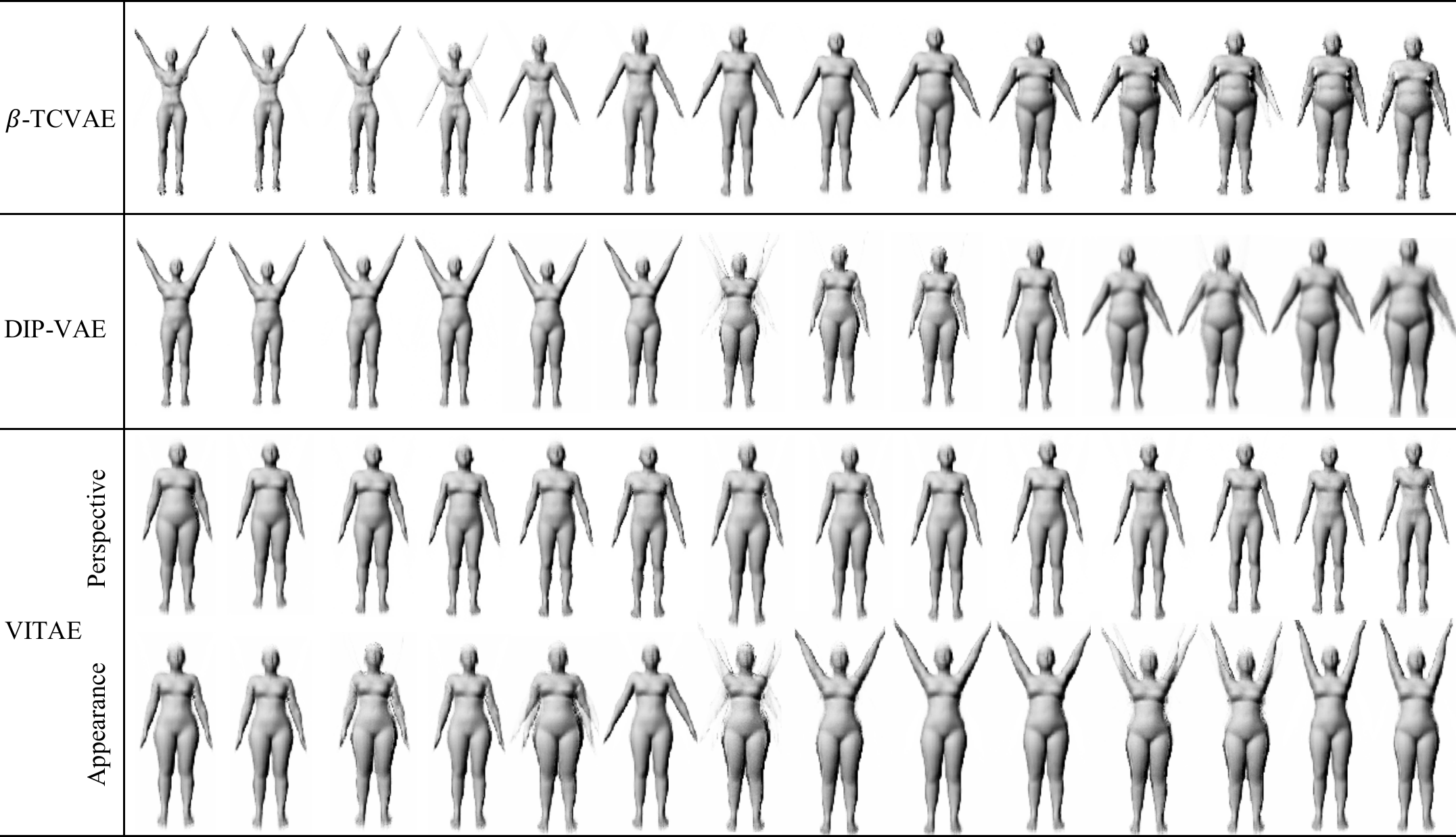}
\caption{Disentanglement of body shape and body pose on SMPL-generated bodies for three different models. The images are generated by varying one latent dimension, while keeping the rest fixed. For the C-VITAE model we have shown this for both the appearance and perspective spaces, since this is the only model where we quantitatively observe disentanglement.}
\label{fig:body_res}
\end{figure}

\textbf{Qualitative evaluation.}
Again, we manipulate the latent codes to visualize their effect (\FIG\ref{fig:body_res}). This time, we here show the result for $\beta$-TCVAE, DIP-VAE-II and VITAE. The results from standard VAE and $\beta$-VAE can be found in \SUPMAT. For both $\beta$-TCVAE and DIP-VAE-II we do not observe disentanglement of body pose and shape, since the decoded images both change arm position (from up to down) and body shape. We note that for both $\beta$-VAE, $\beta$-TCVAE and DIP-VAE-II we did a grid search for their respective hyper parameters. For these three models, we observe that the choice of hyper parameters (scaling of KL term) can have detrimental impact of reconstructions and generated samples. Due to lack of space, test set reconstructions and generated samples can be found in the \SUPMAT. For VITAE we observe some disentanglement of body pose and shape, as variation in appearance space mostly changes the positions of the arms, while the variations in the perspective space mostly changes body shape. The fact that we cannot achieve full disentanglement of this SMPL dataset indicates the difficulty of the task.

\subsection{Disentanglement on CelebA}

\begin{figure}[h!]
\centering
\subfloat[Changing $\z_{P,1}$ corresponds to facial size.]{
	\includegraphics[width=0.11\textwidth]{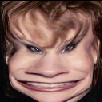} \hspace{-0.17cm}
	\includegraphics[width=0.11\textwidth]{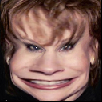} \hspace{-0.17cm}
	\includegraphics[width=0.11\textwidth]{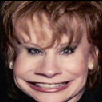} \hspace{-0.17cm}
	\includegraphics[width=0.11\textwidth]{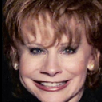} \hspace{-0.17cm}
	\includegraphics[width=0.11\textwidth]{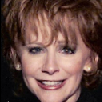} \hspace{-0.17cm}
	\includegraphics[width=0.11\textwidth]{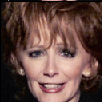} \hspace{-0.17cm}
	\includegraphics[width=0.11\textwidth]{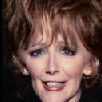} \hspace{-0.17cm}
	\includegraphics[width=0.11\textwidth]{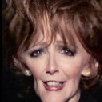} \hspace{-0.17cm}
	\includegraphics[width=0.11\textwidth]{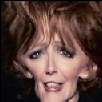} \hspace{-0.17cm}
} \\ \vspace{-0.1cm}
\subfloat[Changing $\z_{P,2}$ corresponds to facial displacement.]{
	\includegraphics[width=0.11\textwidth]{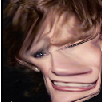} \hspace{-0.17cm}
	\includegraphics[width=0.11\textwidth]{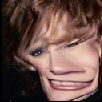} \hspace{-0.17cm}
	\includegraphics[width=0.11\textwidth]{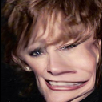} \hspace{-0.17cm}
	\includegraphics[width=0.11\textwidth]{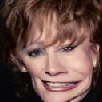} \hspace{-0.17cm}
	\includegraphics[width=0.11\textwidth]{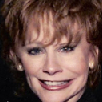} \hspace{-0.17cm}
	\includegraphics[width=0.11\textwidth]{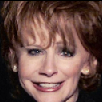} \hspace{-0.17cm}
	\includegraphics[width=0.11\textwidth]{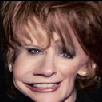} \hspace{-0.17cm}
	\includegraphics[width=0.11\textwidth]{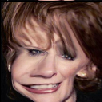} \hspace{-0.17cm}
	\includegraphics[width=0.11\textwidth]{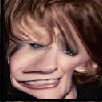} \hspace{-0.17cm}
} \\ \vspace{-0.1cm}
\subfloat[Changing $\z_{A,2}$ corresponds to hair color.]{
	\includegraphics[width=0.11\textwidth]{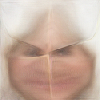} \hspace{-0.17cm}
	\includegraphics[width=0.11\textwidth]{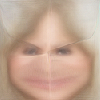} \hspace{-0.17cm}
	\includegraphics[width=0.11\textwidth]{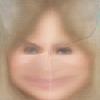} \hspace{-0.17cm}
	\includegraphics[width=0.11\textwidth]{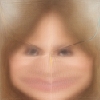} \hspace{-0.17cm}
	\includegraphics[width=0.11\textwidth]{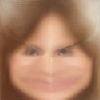} \hspace{-0.17cm}
	\includegraphics[width=0.11\textwidth]{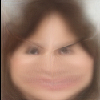} \hspace{-0.17cm}
	\includegraphics[width=0.11\textwidth]{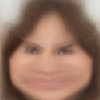} \hspace{-0.17cm}
	\includegraphics[width=0.11\textwidth]{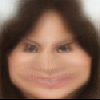} \hspace{-0.17cm}
	\includegraphics[width=0.11\textwidth]{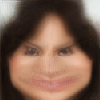} \hspace{-0.17cm}
} \\
\caption{Traversal in latent space shows, that our model can disentangle complex factors such as facial size, facial position and hair color.}
\label{fig:celebA}
\end{figure}

Finally, we qualitatively evaluated our proposed model on the CelebA dataset \cite{liu2015faceattributes}. Since this is a " real life " dataset we do not have access to generative factors and we can therefore only qualitatively evaluate the model. We again model the perspective factors using the aforementioned CPAB transformations, which we assume can model the facial shape. The results can be seen in \FIG\ref{fig:celebA}, which shows latent traversals of both the perspective and appearance factors, and how they influence the generated images. We do observe some interpolation artifacts that are common for architectures using spatial transformers. 

\section{Summary}
In this paper, we have shown how to explicitly disentangle \emph{appearance} from
\emph{perspective} in a variational autoencoder \cite{auto_encoding_variational_bayes, stochastic_backpropagation_and_approximate_inference}.
This is achieved by incorporating a spatial transformer layer \cite{spatial_transformer_nets}
into both encoder and decoder in a coupled manner.
The concepts of appearance and perspective are broad as is evident from
our experimental results in human body images, where they correspond to \emph{pose} and
\emph{shape}, respectively.
By choosing the class of transformations in accordance with prior knowledge
it becomes an effective tool for controlling the inductive bias needed for
disentangled representation learning. On both MNIST and body images our method quantitatively and qualitatively outperforms general purpose disentanglement models \cite{beta_vae, Chen2018, Kumar2017}. 
We find it unsurprisingly that in situations where some prior knowledge about the generative factors is known, encoding these in the into the model give better result than ignoring such information.
%On both MNIST and body images our method quantitatively and qualitatively
%outperforms $\beta$-VAE \cite{beta_vae}, which is the current state-of-the-art.

Our results support the hypothesis \cite{towards_a_definition_of_disentangled_representation} that
inductive biases are necessary for learning disentangled representations, and our
model is a step in the direction of getting fully disentangled generative models.
We envision that the VITAE model should be combined with other models, by first
using the VITAE model to separate appearance and perspective, and then training
a second model only on the appearance. This will factor out one latent
factor at a time, leaving a hierachy of disentangled factors.

%\paragraph{Acknowledgements.} 
%{%\small
%This project has received funding from the European Research Council (ERC) under the European Union's Horizon 2020 research and innovation programme (grant agreement n\textsuperscript{o} 757360). NSD and SH were supported in part by a research grant (15334) from VILLUM FONDEN.}

\paragraph{Acknowledgements.} 
{%\small
This project has received funding from the European Research Council (ERC) under the European Union's Horizon 2020 research and innovation programme (grant agreement n\textsuperscript{o} 757360). NSD and SH were supported in part by a research grant (15334) from VILLUM FONDEN. We gratefully acknowledge the support of NVIDIA Corporation with the donation of GPU hardware used for this research.
}

%%%%%%%%%%% BIBTEX %%%%%%%%%%%
{\small
\bibliographystyle{abbrvnat}
\bibliography{references}
}

\begin{appendix}
\section{Derivation of the ELBO for C-VITAE and U-VITAE}
\label{app:vitae_elbo_derivation}
We will here focus on deriving the ELBO for the C-VITAE, because as we will see the ELBO for the U-VITAE can easily be identified from this. For both models it hold that the generative model is given by

\begin{equation} \nonumber
\label{eq:genrative_model}
p(\x) = \iint p(\x | \z_A, \z_P ) p(\z_A) p(\z_P) \mathrm{d}\z_A \mathrm{d}\z_P.
\end{equation}

We know assume that the inference of appearance now becomes dependent on the perspective factors $\z_P$ \ie

\begin{equation} \nonumber
\label{eq:vitae_ci} p(\z_P | \x) \approx q_P(\z_P | \x) \: \text{and} \: p(\z_A | \x) \approx q_A(\z_A | \x, \z_P) .
\end{equation}

as in the C-VITAE model. The log-posterior is then given by:

\begin{align*}
\log p(\x) &= \log \left( \iint p(\x | \z_A, \z_P) p(\z_A) p(\z_P) \mathrm{d}\z_A \mathrm{d}\z_P \right) \\
&= \log \left( \iint p(\x | \z_A, \z_P) p(\z_A) p(\z_P) \dfrac{q_A(\z_A | \z_P, \x)}{q_A(\z_A | \z_P, \x)} \dfrac{q_P(\z_P | \x)}{q_P(\z_P | \x)} \mathrm{d}\z_A \mathrm{d}\z_P \right) \\
&= \log \left( \int \mathbb{E}_{q_A(\z_A | \z_P, \x)}\left[ \dfrac{p(\x | \z_P, \z_A) p(\z_A)}{q_A(\z_A | \z_P, \x)} \right] p(\z_P) \dfrac{q_P(\z_P | \x)}{q_P(\z_P | \x)} \mathrm{d}\z_P \right) \\
&=\log \mathbb{E}_{q_P(\z_P | \x)} \left[ \mathbb{E}_{q_A(\z_A | \z_P, \x)}\left[ \dfrac{p(\x | \z_P, \z_A) p(\z_A)}{q_A(\z_A | \z_P, \x)} \right] \dfrac{p(\z_P)}{q_P(\z_P |\x)} \right] \\
\end{align*}

By using Jensen's inequality once to exchange the outer expectation with the $\log$ gives us

\begin{align*}
\log p(\x) &\geq \mathbb{E}_{q_P(\z_P | \x)} \left[ \log \left( \mathbb{E}_{q_A(\z_A | \z_P, \x)}\left[ \dfrac{p(\x | \z_P, \z_A) p(\z_A)}{q_A(\z_A | \z_P, \x)} \right] \right) + \log \left(\dfrac{p(\z_P)}{q_P(\z_P |\x)} \right) \right] \\
&=\mathbb{E}_{q_P(\z_P | \x)} \left[ \log \left( \mathbb{E}_{q_A(\z_A | \z_P, \x)}\left[ \dfrac{p(\x | \z_P, \z_A) p(\z_A)}{q_A(\z_A | \z_P, \x)} \right] \right) \right] - D_{KL}(q_P(\z_P | \x) || p(\z_P)) 
\end{align*}

Then, by using Jensen's inequality once more to exchange the $\log$ and inner expectation we get
{\small
\begin{align*}
\log p(\x) &\geq \mathbb{E}_{q_P(\z_P | \x)} \left[ \mathbb{E}_{q_A(\z_A | \z_P, \x)}\left[ \log p(\x | \z_P, \z_A) + \log \left(\dfrac{p(\z_A)}{q_A(\z_A | \z_P, \x)} \right] \right) \right] - D_{KL}(q_P(\z_P | \x) || p(\z_P))  \\
&=\underbrace{\mathbb{E}_{q_P(\z_P | \x)} \left[ \mathbb{E}_{q_A(\z_A | \z_P, \x)}\left[ \log p(\x | \z_P, \z_A)\right] \right]}_\text{term 1} - \underbrace{\mathbb{E}_{q_P(\z_P | \x)} \left[D_{KL}(q_A(\z_A | \z_P, \x) || p(\z_A)) \right]}_\text{term 2} - \underbrace{D_{KL}(q_P(\z_P | \x) || p(\z_P))}_\text{term 3}
\end{align*}
}
Here term 1 is reconstruction term between $\x$ and $p(\x | \z_A , \z_P)$, is the term 2 is the KL divergence for the appearance space $q_A(\z_A | \z_P, \x)$ and its prior $p(\z_A)$ and term 3 is the KL divergence for the perspective space $q_P(\z_P | x)$ and its prior $p(\z_P)$. Similar to how gradients are calculate in VAE's, the outer expectation in term 2 is calculated with respect to a single sample, but can also be computed with respect to multiple samples similar to the work of \citet{importance_weighted_autoencoders}.

To get the ELBO of the U-VITAE model, we make the the inference of the latent spaces independent of each other \ie $q_A(\z_A | z_P, x) = q_A(\z_A | x)$. This get rid of the expectation in term 2 and we are left with

\begin{equation} \nonumber
\log p(\x) \geq \mathbb{E}_{q_P(\z_P | \x)} \left[ \mathbb{E}_{q_A(\z_A | \z_P, \x)}\left[ \log p(\x | \z_P, \z_A)\right] \right] - D_{KL}(q_A(\z_A | \x) || p(\z_A)) - D_{KL}(q_P(\z_P | \x) || p(\z_P)),
\end{equation}
which is the ELBO for the U-VITAE model. The intuition behind this equation is that the U-VITAE model is just a standard VAE, where the latent space $\z$ has been split into two smaller latent spaces $\z_P, \z_A$, thus this is reflected in ELBO where the KL-term is similar split into two terms.

\section{Implementation details for the experiments}
\label{app:implementation_notes}

Below we describe the network architectures in details. All models were trained using the Adam optimizer \cite{adam_optimizer} with fixed learning rate of $10^{-4}$. For the MNIST experiments we used a batch size of 512 and trained for a 2000 epochs and for SMLP and CelebA experiments we used a batch size of 256 and trained for 5000 epochs. No early stopping was used. Similar to \citet{ladder_variational_autoencoders}, we use annealing/warmup for the KL-divergence by scaling the term(s) by $w=\min\left(\frac{\text{epoch}}{\text{warmup}}, 1 \right)$, where the warmup parameter was set to half the number of epochs. 

\textbf{Details for MNIST experiments}. Pixel values of the images are scaled to the interval [0,1]. Each pixel is assumed to be Bernoulli distributed. For the encoders and decoders we use multilayer perceptron networks. For the VAE, $\beta$-VAE \cite{beta_vae}, $\beta$-TCVAE \cite{Chen2018} and DIP-VAE \cite{Kumar2017}, we use the settings listed below. For both VITAE models, we model both encoders and both decoders with approximately half the neurons, for a fair comparison. In practice we found that the encoders/decoders of the appearance factors benefits from having a bit higher capacity than the encoders/decoders of the perspective factors.

\begin{table}[h!]
\centering
\begin{tabular}{l|c|c|c}
					 	& Layer 1 		& Layer 2 		& Layer 3 \\ \hline
$\boldsymbol{\mu}_{encoder}$		& 128, (LeakyReLU)		& 64, (LeakyReLU)		& d, (Linear) \\
$\boldsymbol{\sigma}^2_{encoder}$ & 128, (LeakyReLU) & 64, (LeakyReLU) & d, (softplus) \\
$\boldsymbol{\mu}_{decoder}$		& 64, (LeakyReLU)		& 64, (LeakyReLU)		& D, (Sigmoid) \\
\end{tabular}
\caption{Model architecture for the MNIST experiments.}
\label{tab:app:mnist_net}
\end{table}

Here $D=784$ and $d=4$ for VAE based models and $d=2$ for VITAE based models. The numbers corresponds to the size of the layer and the parenthesis is the used activation function. For the LeakyRelu activation function we use hyper parameter $\alpha=0.1$. We only parametrize a mean function in the decoder because we assume the output pixels are Bernoulli distributed. 

\textbf{Details for SMPL experiments}. Images was generated using the SMPL library\footnote{\url{http://smpl.is.tue.mpg.de/}}. The parameters for generating the body shape was drawn from a $\Ncal(0, 1.25^2)$ distribution. The parameters that controls the body pose was uniformly sampled from one out of 4 pre-specified pose configurations, see \TAB\ref{tab:pose}. 

\begin{table}[h!]
\centering
\begin{tabular}{l|c|c|c|c}
				& Pose 1		& Pose 2		& Pose 3		& Pose 4 \\ \hline 
Left shoulder	& $-\pi/8$	& $-\pi/16$	& $\pi/16$	& $\pi/8$ \\ \hline
Right shoulder	& $\pi/8$	& $\pi/16$		& $-\pi/16$	& $-\pi/8$ \\ \hline
Left arm			& $-\pi/3.5$	& $-\pi/3.5$	& $\pi/3.5$	& $\pi/3.5$ \\ \hline
Right arm		& $\pi/3.5$	& $\pi/3.5$	& $-\pi/3.5$	& $-\pi/3.5$ 
\end{tabular}
\caption{When generating synthetic bodies, we uniformly sample one of the above settings for the pose.}
\label{tab:pose}
\end{table}

The resolution of each image was scaled down to $(400, 200)$. Each pixel is assumed to be Normal distributed. For the VAE based models, we use the settings listed below. For the VITAE models we used approximately half the neurons for the encoders/decoders. 

\begin{table}[h!]
\centering
\begin{tabular}{l|c|c|c}
					 	& Layer 1 		& Layer 2 		& Layer 3 \\ \hline
$\boldsymbol{\mu}_{encoder}$		& 256, (LeakyReLU)		& 128, (LeakyReLU)		& d, (Linear) \\
$\boldsymbol{\sigma}^2_{encoder}$ & 256, (LeakyReLU) & 128, (LeakyReLU) & d, (softplus) \\
$\boldsymbol{\mu}_{decoder}$		& 128, (LeakyReLU) & 256, (LeakyReLU)	& D, (Linear) \\
\end{tabular}
\caption{Model architecture for the SMPL experiments.}
\label{tab:app:smpl_net}
\end{table}

Here $D=80.000$ and $d=4$ for VAE based models and $d=2$ for VITAE based models. The numbers corresponds to the size of the layer and the parenthesis is the used activation function. For the LeakyRelu activation function we use hyper parameter $\alpha=0.1$. We only parametrize a mean in the decoder because the variance function is in general very hard to train and completely arbitrarily outside the latent manifold \cite{latent_space_oddity}. It was therefore fixed for all pixels in all images to $\sigma^2_{decoder}=0.1$. For the CPAB transformations \cite{highly_expressive_spaces_of_well_behaved_transformations} we ran the experiments with tessellation parameters [2, 4] with zero boundary constrains and no volume preservation constrains. With these settings, we are generating perspective transformations of size 30 \ie $\text{dim}(\btheta)=30$. 

\textbf{Details for CelebA experiments}. We use the align and cropped version of the dataset, downloaded from the homepage\footnote{\url{http://mmlab.ie.cuhk.edu.hk/projects/CelebA.html}}. Each image was then down sampled to size $128\times 128$, to decrease computational time. Each pixel is assumed to be Normal distributed. For this task we use a convolutional-VAE. Below is listed the configuration of the network:

\begin{table}[h!]
\centering
\resizebox{\textwidth}{!}{
\setlength\tabcolsep{3pt}
\begin{tabular}{l|c|c|c|c}
					 				& Layer 1 						& Layer 2 						& Layer 3 				& Layer 4\\ \hline
$\boldsymbol{\mu}_{encoder}$		& Conv(10, 5, 2, LeakyReLU)		& Conv(20, 5, 2, LeakyReLU)		& Conv(40, 3, 2, LeakyReLU)		& Dense(2, Linear) \\
$\boldsymbol{\sigma}^2_{encoder}$ 	& Conv(10, 5, 2, LeakyReLU) 	& Conv(20, 5, 2, LeakyReLU)		& Conv(40, 3, 2, LeakyReLU)		& Dense(2, Softplus) \\
$\boldsymbol{\mu}_{decoder}$		& DeConv(40, 3, 2, LeakyReLU) 	& DeConv(20, 3, 2, LeakyReLU)	& DeConv(10, 5, 2, LeakyReLU)	& DeConv(3, 5, 2, Sigmoid) \\
\end{tabular}
}
\caption{Model architecture for the CelebA experiments. Conv denotes a convolutional layer and DeConv denotes de-/transposed convolutional layers. The parameters are respective number of filters, filter size, stride and activation function.}
\label{tab:app:smpl_net}
\end{table}

For the CPAB transformation \cite{highly_expressive_spaces_of_well_behaved_transformations} we ran the experiments with tessellation parameters [4, 4] with zero boundary constrains and no volume preservation constrains. With these settings, we are generating perspective transformations of size 62. 

\textbf{Computational requirements.}
Even though VITAE has a more complicated architecture than VAE
(comparing \FIG(3a) vs. (3c) in main paper) both forward and backward
passes in the models have roughly the same complexity when we use affine transformations
(see Table~\ref{tab:timings}). Using the more complex CPAB transformations adds
some penalty to the computational time.

\begin{table}[h!]
\centering

\begin{tabular}{ll|r|r|}
                         &        & Forward & Backward \\ \hline
\multicolumn{2}{l|}{VAE \& $\beta$-VAE }          & 0.0016s  & 0.014s    \\ \hline
\multicolumn{2}{l|}{$\beta$-TCVAE} & 0.0020s & 0.016s \\ \hline
\multicolumn{2}{l|}{DIP-VAE-II} 	& 0.0025s & 0.018s \\ \hline
\multirow{2}{*}{C-VITAE} & Affine & 0.0092s  & 0.037s    \\ \cline{2-4} 
                         & CPAB   & 0.1s   	& 0.86s    \\ \hline
\end{tabular}
\vspace{0.1cm}
\caption{Forward and backwards timings for the different architectures. The experiments was conducted with an Intel Xeon E5-2620v4 CPU and Nvidia GTX TITAN X GPU.}
\label{tab:timings}
\end{table}
\section{Stability results}
\begin{figure}[h]
\centering
\includegraphics[width=1.0\textwidth, trim=0.4in 3.3in 0.45in 2.95in, clip]{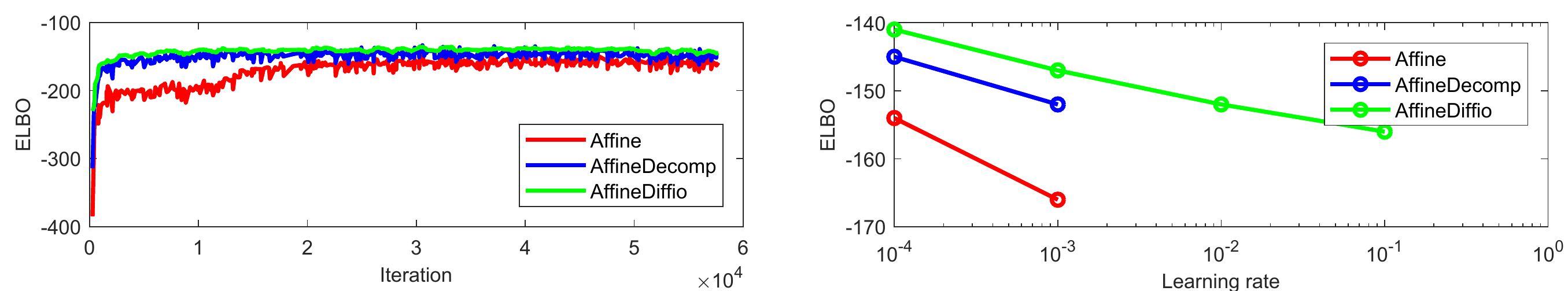}
\caption{Top: Stability towards choice of learning rate for three different parametrizations of affine transformations. Missing values indicates that the network diverged. Bottom: Learning curves also show that the diffiomorphic affine parametrization converges faster and is more stable in its training. }
\label{fig:stability}
\end{figure}

In the main paper we discuss multiple ways to parameterize an affine transformation. If we choose $\Tcal_{\gamma}$ with a diffiomorphic parameterization, we have found that this also has positive positive optimization properties. \FIG\ref{fig:stability} shows the ELBO as a function of the learning rate $\lambda$ for the three different choices of affine parametrization discussed in the main paper, using our C-VITAE architecture. We clearly see that the diffeomorphic affine parametrization archives a tighter bound, and can run for much higher learning rates (faster convergence) before the network begins to diverge. These findings are similar to those of \citet{deep_diffeomorphic_transformer_networks} in the supervised context.

These experiments was conducted on the MNIST dataset. For all three experiments we use the C-VITAE architecture with a neural network structure as Table \ref{tab:app:mnist_net}. A batch size of 512 was used. The results where generated by changing the parametrization of the affine spatial transformer between
\begin{align}
&\text{Affine}			&& \Tcal_\bgamma (\x) = \begin{bmatrix}
\gamma_{11} & \gamma_{12} & \gamma_{13} \\
\gamma_{21} & \gamma_{22} & \gamma_{14}
\end{bmatrix} \begin{bmatrix}
x \\ y \\ 1
\end{bmatrix} \\
&\text{AffineDecomp}		&& \Tcal_\bgamma(\x) = \begin{bmatrix}
\cos(\alpha) & -\sin(\alpha) \\
\sin(\alpha) & \;\;\;\cos(\alpha) 
\end{bmatrix}
\begin{bmatrix}
1 & m\\
0 & 1
\end{bmatrix} \begin{bmatrix}
s_x & 0 \\
0 & s_y
\end{bmatrix} + \begin{bmatrix}
t_x \\ t_y
\end{bmatrix} \\
&\text{AffineDiffio}		&& \Tcal_\bgamma (\x) = \textbf{expm} \left(\begin{bmatrix}
\gamma_{11} & \gamma_{12} & \gamma_{13} \\
\gamma_{21} & \gamma_{22} & \gamma_{14} \\
0 & 0 & 0
\end{bmatrix} \right) \begin{bmatrix}
x \\ y \\ 1
\end{bmatrix}
\end{align}

and by varying the learning rate $\lambda=\{10^{-4}, 10^{-3}, 10^{-2}, 10^{-1}\}$. The lower subplot of Figure 4, was generated using a learning rate of $\lambda=10^{-4}$ to make sure that all transformer types would converge.

\section{Additional results}
\subsection{MNIST experiments}
\begin{figure}[h!]
\centering
\subfloat[VAE]{\includegraphics[width=0.48\textwidth, trim=3cm 0cm 0cm 0cm, clip]{figs/reconstructions_vae.PNG}} \hspace{1mm}
\subfloat[$\beta$-VAE]{\includegraphics[width=0.48\textwidth, trim=3cm 0cm 0cm 0cm, clip]{figs/reconstructions_betavae.PNG}}

\subfloat[$\beta$-TCVAE]{\includegraphics[width=0.48\textwidth, trim=0cm 0cm 0cm 0cm, clip]{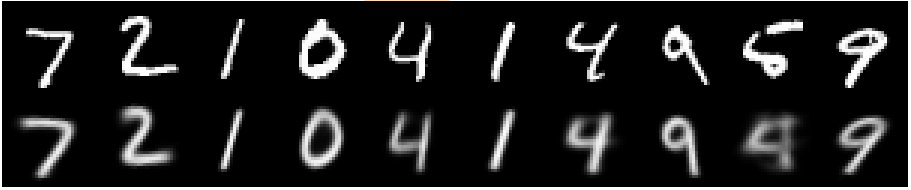}} \hspace{1mm}
\subfloat[DIP-VAE-II]{\includegraphics[width=0.48\textwidth, trim=0cm 0cm 0cm 0cm, clip]{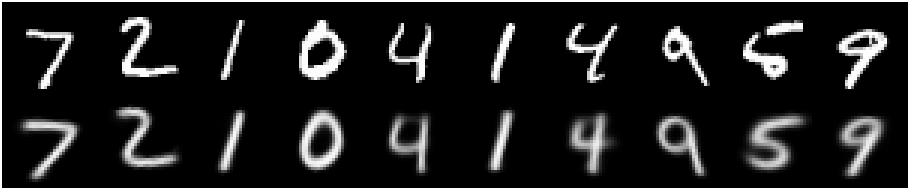}}

\subfloat[C-VITAE]{\includegraphics[width=0.48\textwidth, trim=3cm 0cm 0cm 0cm, clip]{figs/reconstructions_vitae.PNG}}
\caption{Samples from the test set (top rows) and the corresponding reconstructions (bottom rows) for all models. We clearly observe that the additional weight on the KL term in $\beta$-VAE, $\beta$-TCVAE and DIP-VAE-II makes the reconstructions worse.}
\label{fig:mnist_reconstructions}
\end{figure}

\begin{figure}[H]
\centering
\subfloat[VAE]{\includegraphics[width=0.32\textwidth]{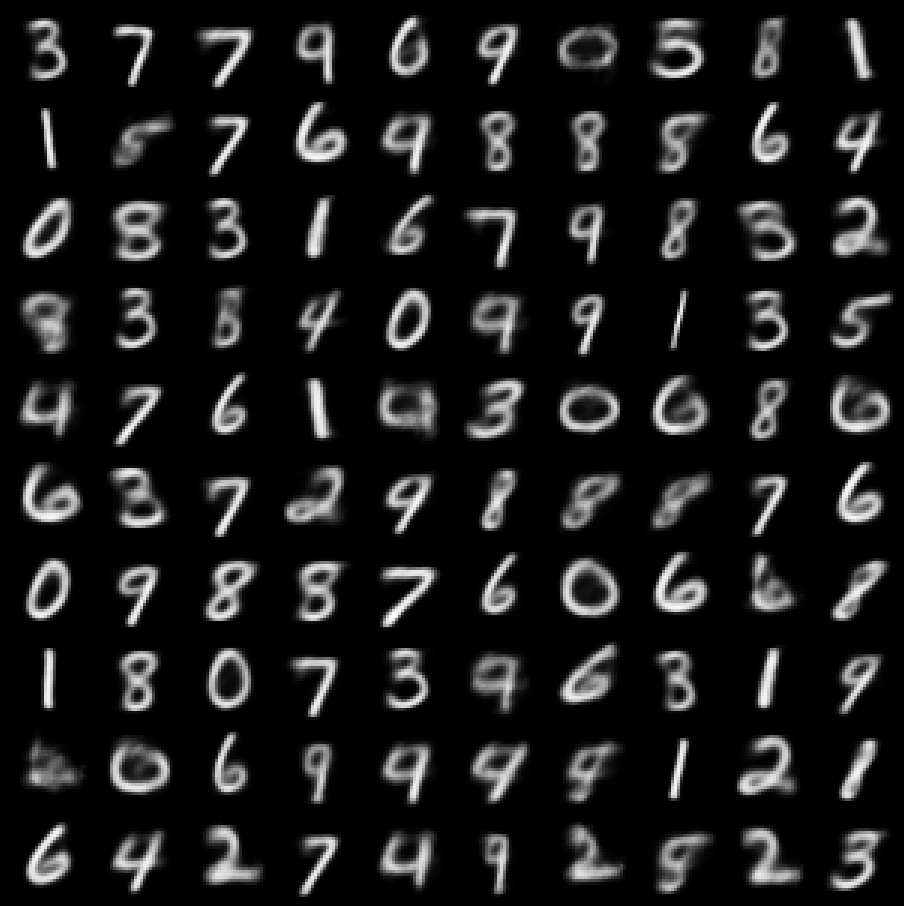}} \hspace{1mm}
\subfloat[$\beta$-VAE]{\includegraphics[width=0.32\textwidth]{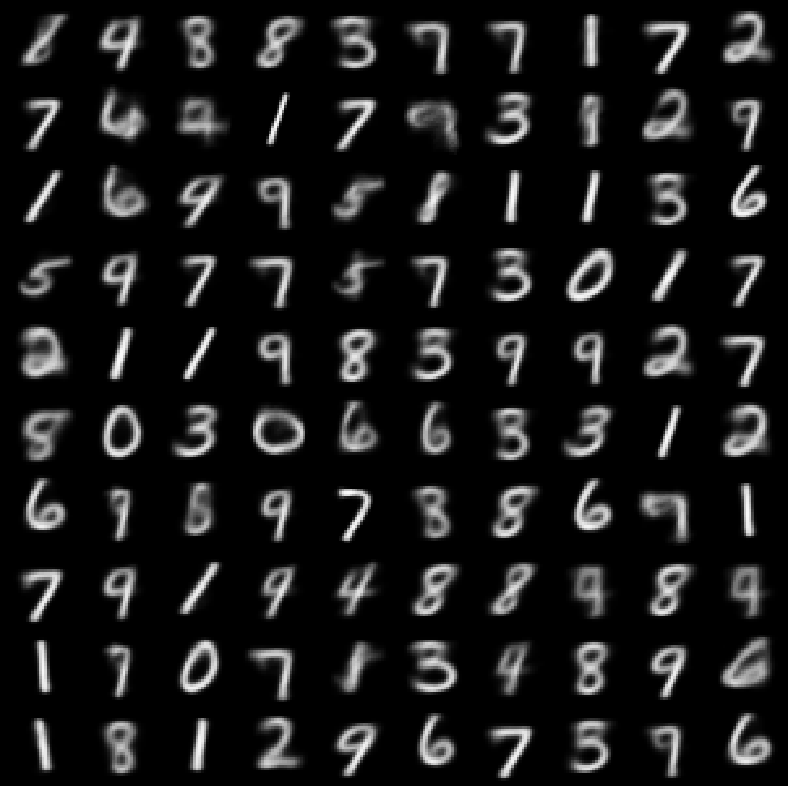}} \hspace{1mm}
\subfloat[$\beta$-TCVAE]{\includegraphics[width=0.32\textwidth]{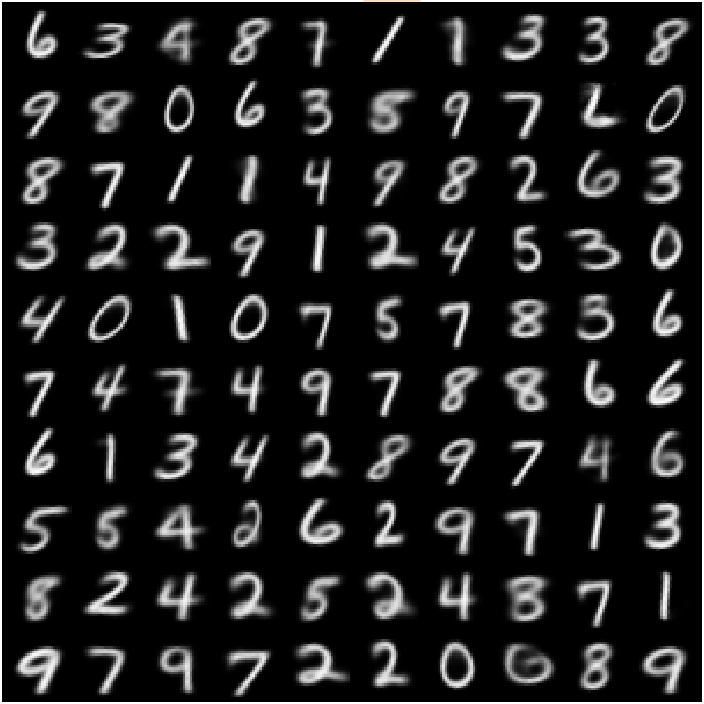}}

\subfloat[DIP-VAE-II]{\includegraphics[width=0.32\textwidth]{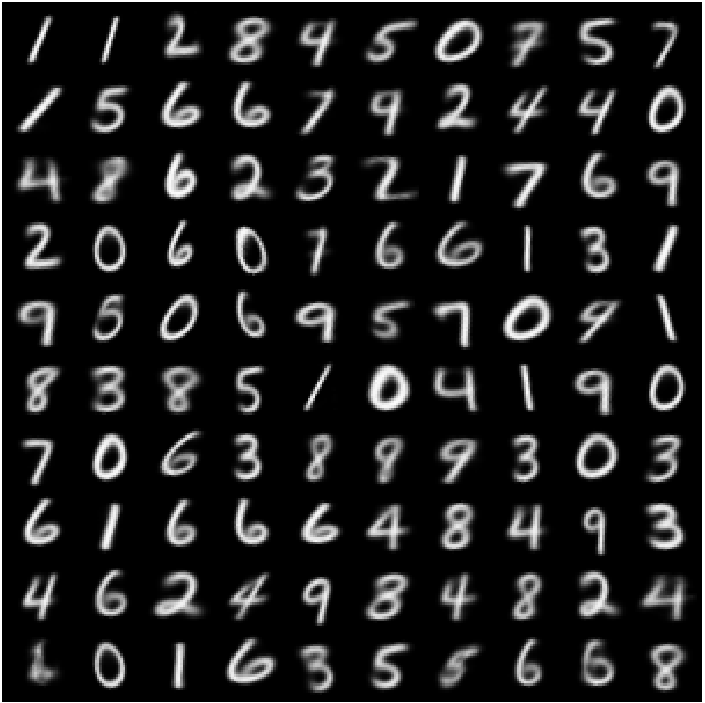}} \hspace{1mm}
\subfloat[C-VITAE]{\includegraphics[width=0.32\textwidth]{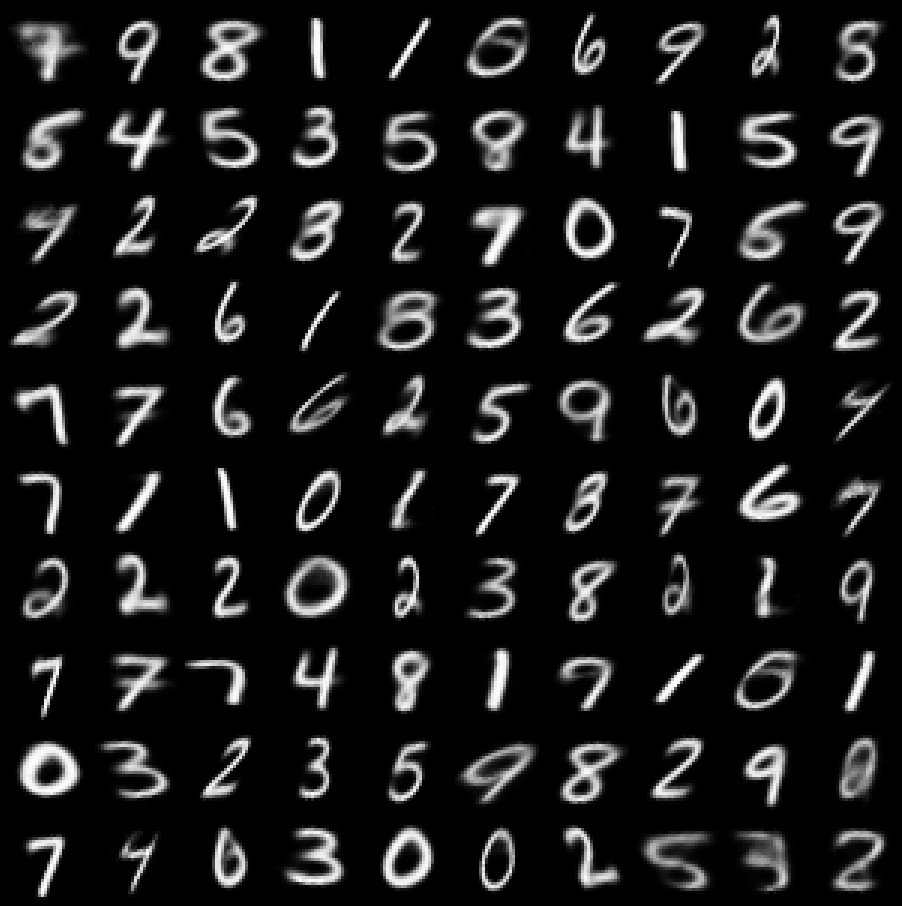}}
\caption{Samples from the prior distribution.}
\label{fig:mnist_samples}
\end{figure}

\begin{figure}
\centering
\subfloat[VAE]{\includegraphics[width=1\textwidth]{figs/interpolation_vae.PNG}}

\subfloat[$\beta$-VAE]{\includegraphics[width=1\textwidth]{figs/interpolation_betavae.PNG}}

\subfloat[$\beta$-TCVAE]{\includegraphics[width=1\textwidth]{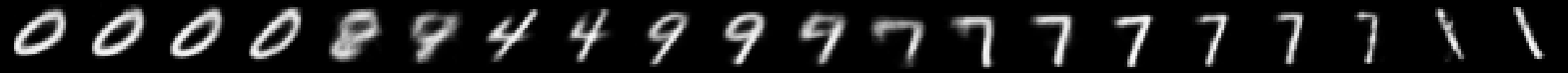}}

\subfloat[DIP-VAE-II]{\includegraphics[width=1\textwidth]{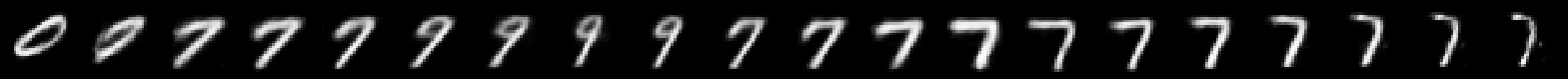}}

\subfloat[C-VITAE]{\includegraphics[width=1\textwidth]{figs/interpolation_vitae.PNG}}
\caption{Latent manipulation. The images were generated by varying one latent dimension, while keeping the rest fixed. We choose the latent variable that qualitatively gave the best results.}
\label{fig:mnist_latent}
\end{figure}

In \FIG\ref{fig:mnist_reconstructions} reconstructions from the different models can be seen. In \FIG\ref{fig:mnist_samples} generated sampler from the different models can be seen. In \FIG\ref{fig:mnist_latent} latent manipulations can be seen.

\section{SMPL experiment}
\begin{figure}[h!]
\centering
\subfloat[VAE]{\includegraphics[width=0.49\textwidth]{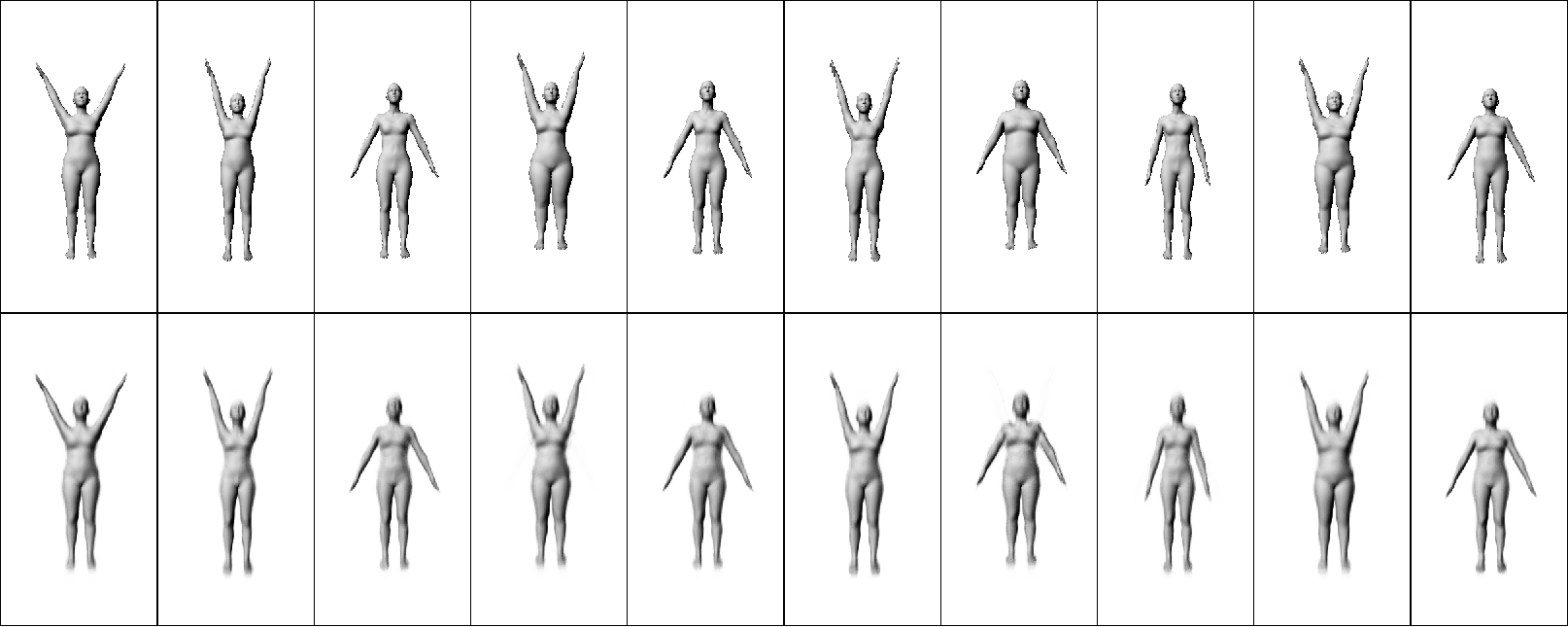}} \hspace{1mm}
\subfloat[$\beta$-VAE]{\includegraphics[width=0.49\textwidth]{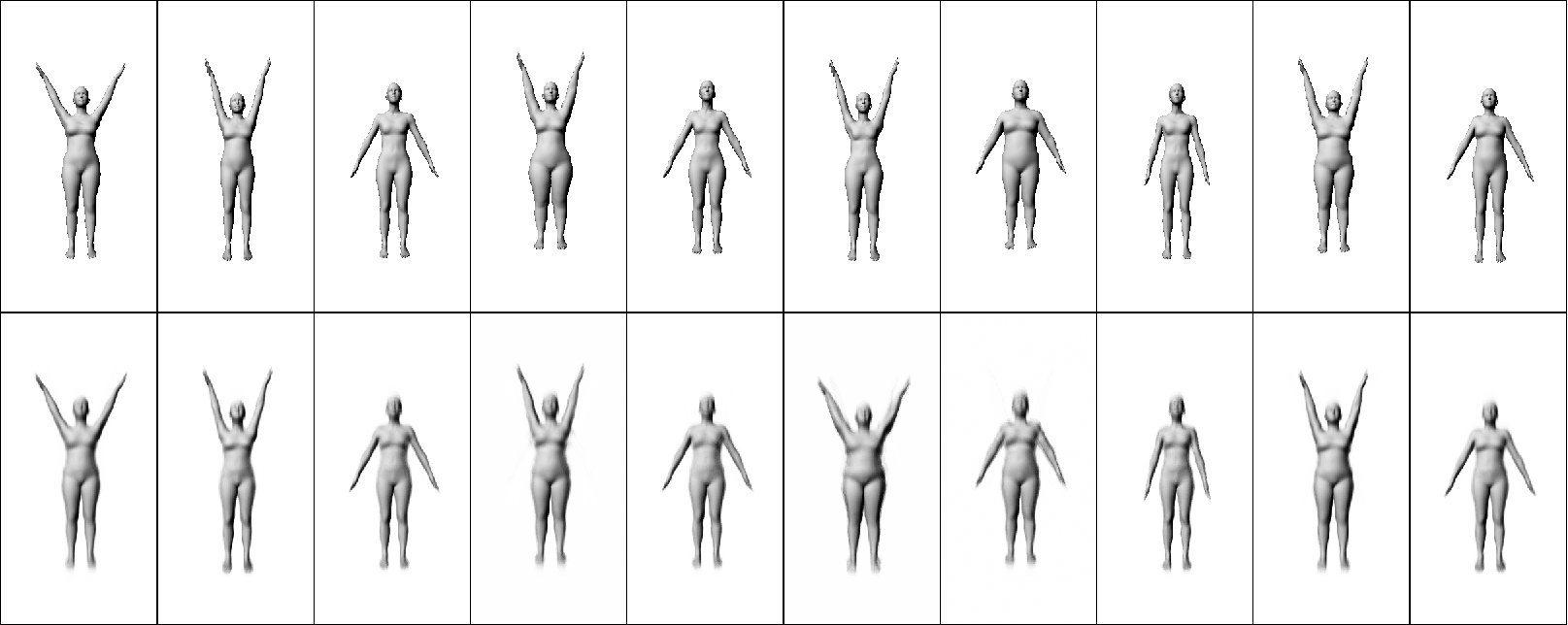}}

\subfloat[$\beta$-TCVAE]{\includegraphics[width=0.49\textwidth]{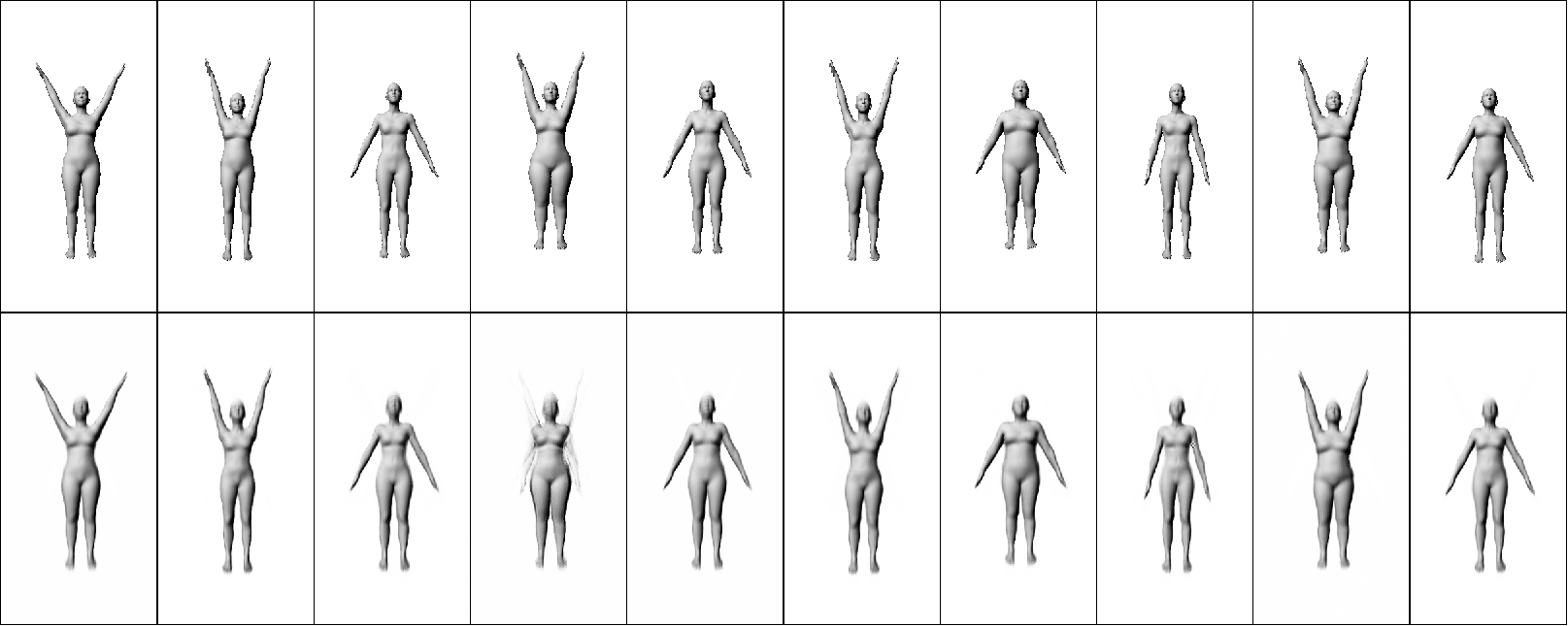}}
\subfloat[DIP-VAE-II]{\includegraphics[width=0.49\textwidth]{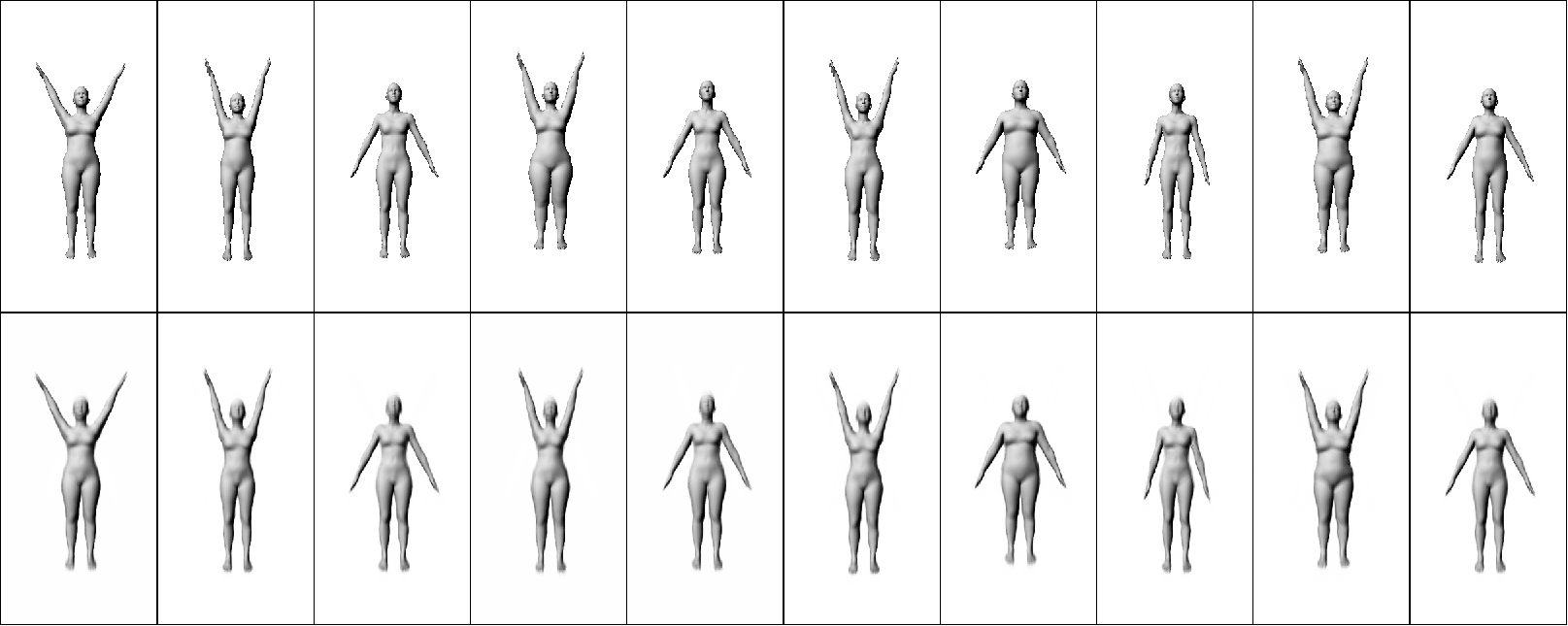}}

\subfloat[C-VITAE]{\includegraphics[width=0.49\textwidth]{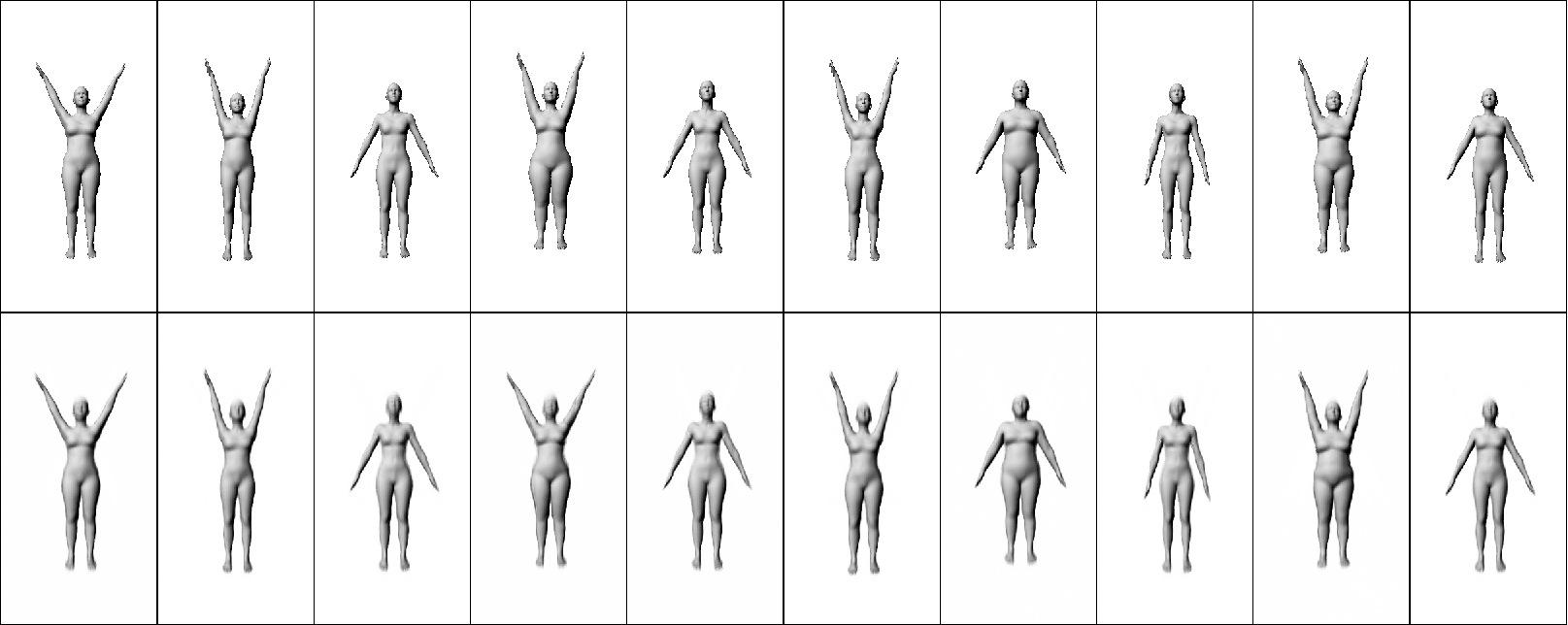}}
\caption{Test set reconstructions on SMPL dataset.}
\label{fig:smpl_reconstructions}
\end{figure}

\begin{figure}[h!]
\centering
\subfloat[VAE]{\includegraphics[width=0.49\textwidth]{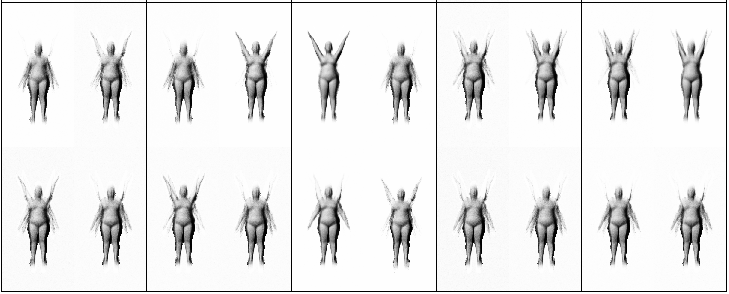}}
\subfloat[$\beta$-VAE]{\includegraphics[width=0.49\textwidth]{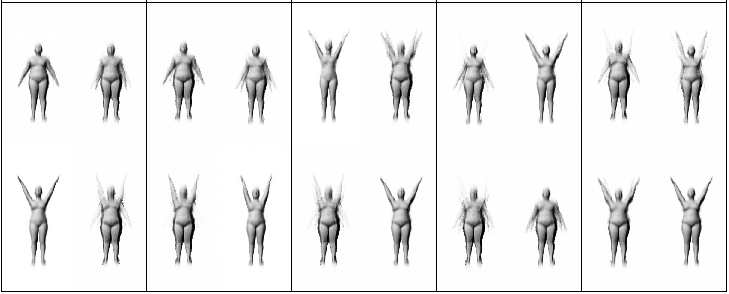}}

\subfloat[$\beta$-TCVAE]{\includegraphics[width=0.49\textwidth]{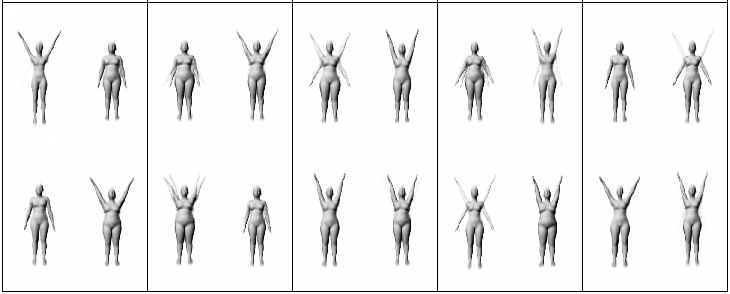}}
\subfloat[DIP-VAE-II]{\includegraphics[width=0.49\textwidth]{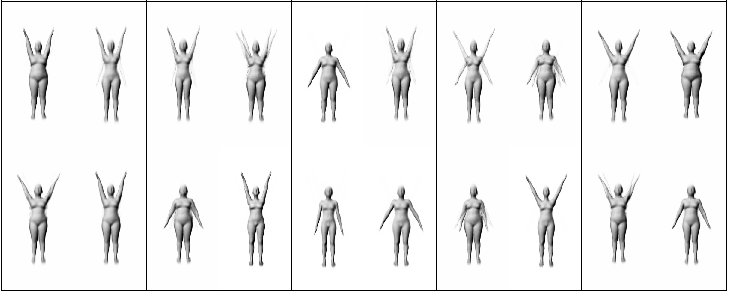}}

\subfloat[C-VITAE]{\includegraphics[width=0.49\textwidth]{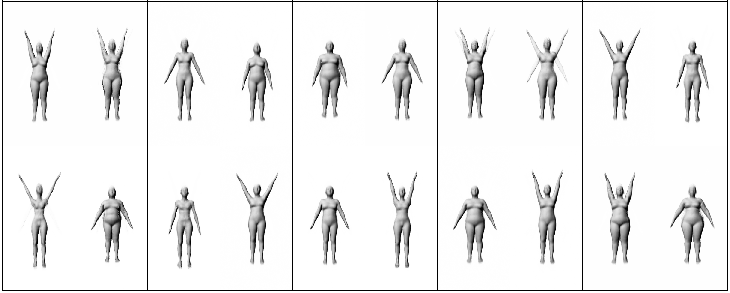}}
\caption{Samples from the prior distribution.}
\label{fig:smpl_samples}
\end{figure}

In \FIG\ref{fig:smpl_reconstructions} reconstructions from the different models can be seen. In \FIG\ref{fig:smpl_samples} generated sampler from the different models can be seen. In \FIG\ref{fig:mnist_latent} latent manipulations can be seen.

\begin{figure}[h!]
\centering
\includegraphics[width=1\textwidth, trim=0cm 0cm 12cm 0cm, clip]{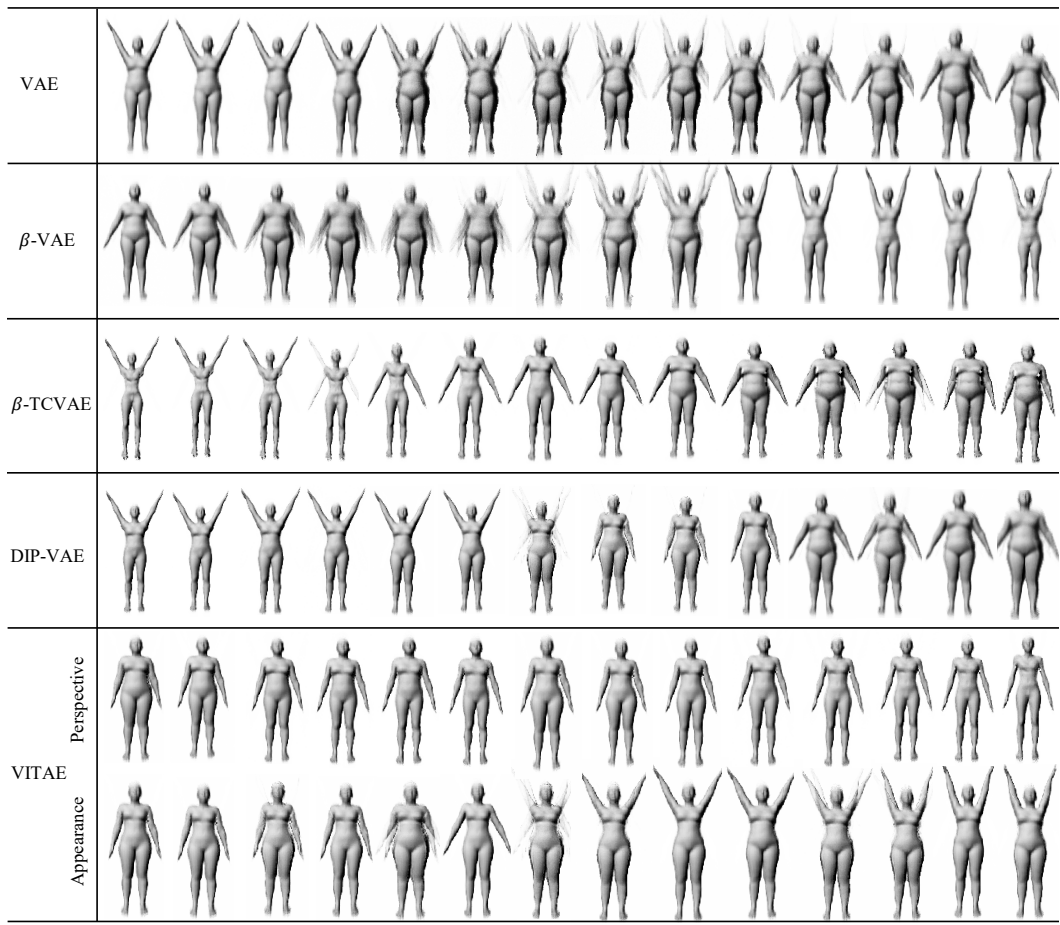}
\caption{Disentanglement of body shape and body pose on SMPL-generated bodies for all models. The images are generated by varying one latent dimension, while keeping the rest fixed. For the C-VITAE model we have shown this for both the appearance and perspective spaces, since this is the only model where we quantitatively observe disentanglement.}
\label{fig:body_res}
\end{figure}

\end{appendix}

\end{document}